\documentclass[preprint,12pt]{elsarticle}

\usepackage[a4paper,top=3cm,bottom=2cm,left=3cm,right=3cm,marginparwidth=1.75cm]{geometry}
\usepackage{amsmath}
\usepackage{multirow}
\usepackage{graphicx}
\usepackage{amssymb}
\usepackage{hyperref}
\usepackage{float}
\usepackage{lscape}
\usepackage{longtable}
\usepackage{array}
\usepackage[T1]{fontenc}
\usepackage{color}
\usepackage[flushleft]{threeparttable}
\usepackage{subcaption}

\begin{document}

\begin{frontmatter}

\title{Generative Technology for Human Emotion Recognition: A Scope Review}

\author[address1]{Fei Ma\fnref{equal}}
\ead{mafei@gml.ac.cn}
\author[address2]{Yucheng Yuan\fnref{equal}}
\ead{yuan-yc23@mails.tsinghua.edu.cn}
\author[address1]{Yifan Xie}
\ead{xieyifan@stu.xjtu.edu.cn}
\author[address3]{Hongwei Ren}
\ead{hren066@connect.hkust-gz.edu.cn}
\author[address4]{Ivan Liu}
\ead{ivanliu@bnu.edu.cn}
\author[address41]{Ying He}
\ead{heying@szu.edu.cn}
\author[address5]{Fuji Ren}
\ead{renfuji@uestc.edu.cn}
\author[address1]{Fei Richard Yu\corref{corresponding}}
\ead{yufei@gml.ac.cn}
\author[address2]{Shiguang Ni\corref{corresponding}}
\ead{ni.shiguang@sz.tsinghua.edu.cn}

\address[address1]{Guangdong Laboratory of Artificial Intelligence and Digital Economy (SZ), Shenzhen, China}
\address[address2]{Tsinghua Shenzhen International Graduate School, Tsinghua University, Shenzhen, China}
\address[address3]{MICS Thrust,
The Hong Kong University of Science and Technology (GZ), Guangzhou, China}
\address[address4]{Department of Psychology, Faculty of Arts and Sciences, Beijing Normal University at Zhuhai, Zhuhai, China}
\address[address41]{College of Computer Science and Software Engineering, Shenzhen University, Shenzhen, China}
\address[address5]{School of Computer Science and Engineering, University of Electronic Science and Technology of China, Chengdu, China}

\fntext[equal]{The two authors contribute equally to this work.}
\cortext[corresponding]{Corresponding author.}

\begin{abstract}
Affective computing stands at the forefront of artificial intelligence (AI), seeking to imbue machines with the ability to comprehend and respond to human emotions. 
Central to this field is emotion recognition, which endeavors to identify and interpret human emotional states from different modalities, such as speech, facial images, text, and physiological signals.
In recent years, important progress has been made in generative models, including Autoencoder, Generative Adversarial Network, Diffusion Model, and Large Language Model.
These models, with their powerful data generation capabilities, emerge as pivotal tools in advancing emotion recognition.
However, up to now, there remains a paucity of systematic efforts that review generative technology for emotion recognition.
This survey aims to bridge the gaps in the existing literature by conducting a comprehensive analysis of over 320 research papers until June 2024. 
Specifically, this survey will firstly introduce the mathematical principles of different generative models and the commonly used datasets.
Subsequently, through a taxonomy, it will provide an in-depth analysis of how generative techniques address emotion recognition based on different modalities in several aspects, including data augmentation, feature extraction, semi-supervised learning, cross-domain, etc.
Finally, the review will outline future research directions, emphasizing the potential of generative models to advance the field of emotion recognition and enhance the emotional intelligence of AI systems.

\end{abstract}



\end{frontmatter}


\section{Introduction}
\label{intro}
\quad
In the field of affective computing \cite{picard2000affective,picard2003affective,poria2017review}, 
researchers are actively engaged in developing technologies that can simulate, recognize, and understand human emotions.
The primary objective of these endeavors is to imbue computers with a profound level of emotional perception, thereby facilitating more intelligent and human-like interactive experiences \cite{tao2005affective,calvo2010affect}.
There are extensive applications of affective computing in numerous fields. For example, affective computing can be employed in healthcare to monitor patients' emotional changes, assist in diagnosing mental disorders, and evaluate treatment outcomes \cite{pepa2021automatic,khanna2022affective}. 
In education, it is employed to develop emotionally intelligent educational systems that adjust teaching strategies based on students' emotional states, thereby enhancing learning effectiveness \cite{yadegaridehkordi2019affective,mejbri2022trends}. 
In the field of autonomous driving, affective computing can monitor drivers' emotional conditions and provide timely warnings for risks such as fatigue driving \cite{zepf2020driver,mou2023driver}. 
In marketing, it enables the analysis of consumers' emotional preferences, providing insights for businesses to optimize marketing strategies and customer service \cite{liu2022artificial,gao2023winning}.

As a core component of affective computing, emotion recognition \cite{sebe2005multimodal,wang2022systematic,ezzameli2023emotion,khare2023emotion} focuses on identifying emotional states from the data that conveys human emotions. 
Due to the inherent heterogeneity of human emotional expressions, these data can come from various modalities, such as speech, facial images, text, and physiological signals. 
Notably, speech emotion recognition (SER) identifies the speaker's emotion by extracting features such as pitch, volume, and speech rate \cite{akccay2020speech}. 
Facial emotion recognition (FER) predicts human emotional states by detecting and tracking facial feature points \cite{canal2022survey}. 
Textual emotion recognition (TER) focuses on identifying the emotional content expressed in written language, such as social media posts, customer reviews, or personal messages \cite{deng2021survey}. 
Physiological signals, such as electroencephalogram (EEG), electrocardiogram (ECG), heart rate variability (HRV), electrodermal activity (EDA), has become more prevalent in recent years due to their ability to objectively measure an individual's emotional state \cite{egger2019emotion}. 
In addition, multimodal emotion recognition (MER) also receives increasing attention \cite{baltruvsaitis2018multimodal,ahmed2023systematic}. 
Compared with emotion recognition based on a single modality, it comprehensively utilizes emotional information from different modalities to improve the performance of emotion recognition.

Over the past few decades, 
researchers have conducted extensive work on emotion recognition based on machine learning and deep learning \cite{ezzameli2023emotion,dzedzickis2020human}, among which generative technology-based methods have shown tremendous potential. 
In contrast to discriminative models that directly learn the decision boundaries between different emotion categories, 
generative methods focus on learning the intrinsic distribution and representation of the emotional data \cite{jebara2012machine,jabbar2021survey}. 
Generative models, with their powerful generative capabilities, can generate samples highly similar to real emotional data, effectively enhancing the performance and generalization ability of emotion recognition. 
However, to the best of our knowledge, there is currently a lack of a systematical review that summarizes the work of generative technology for emotion recognition.

\begin{figure}[t]
  \centering
  \includegraphics[width=1\textwidth]{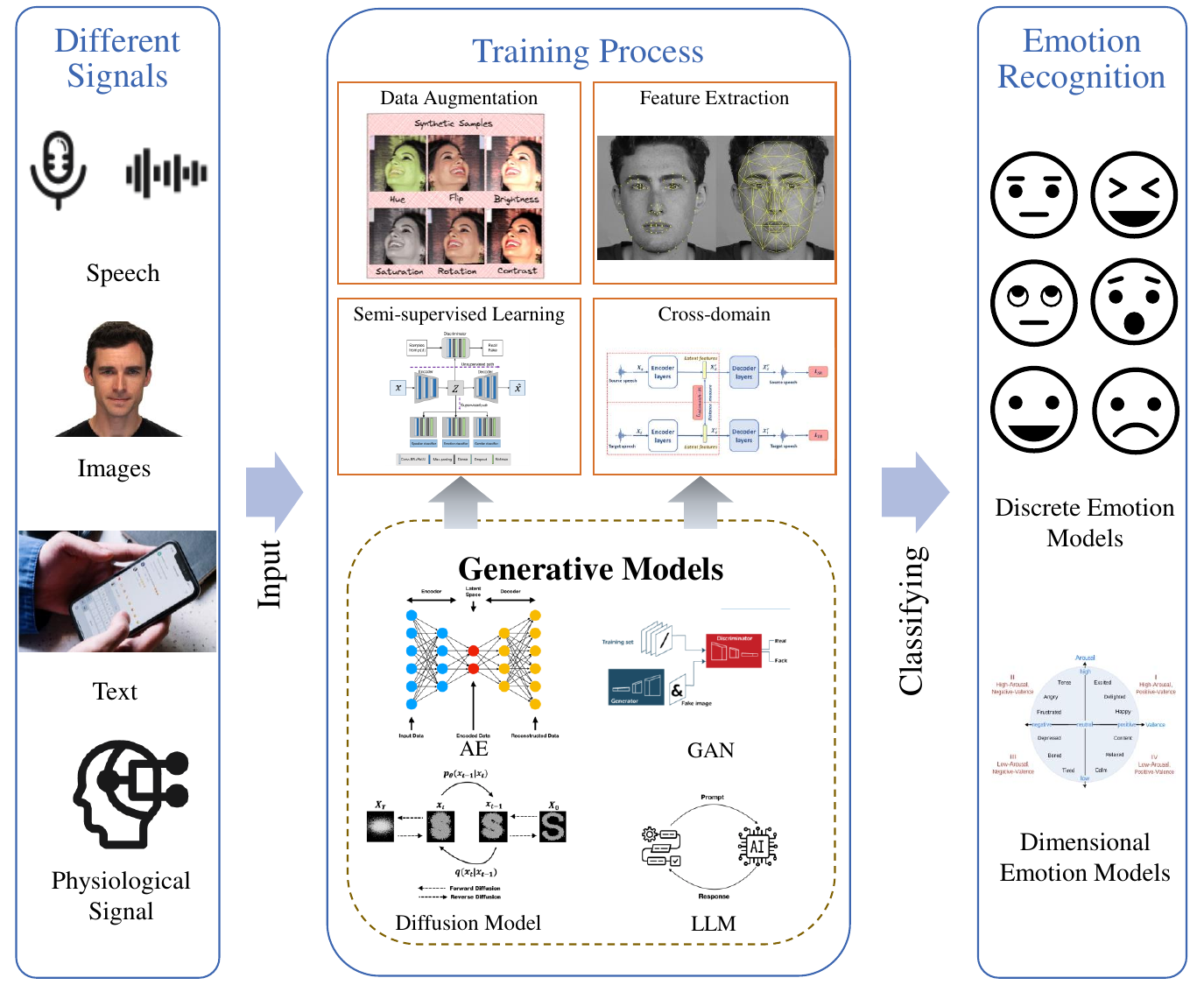}
  \caption{Schematic Diagram of Generation technology for Emotion Recognition. \protect\footnotemark}
  \label{schematic diagram}
\end{figure}

In view of this,  
this paper aims to provide a comprehensive overview of the research progress in generative technology for emotion recognition, based on more than 320 technical papers up to June 2024. The overall flow is shown in Figure \ref{schematic diagram}. 
Specifically, this survey will focus on several representative models, such as Autoencoder (AE) \cite{rumelhart1986learning}, Generative Adversarial Network (GAN) \cite{goodfellow2014generative}, Diffusion Model (DM) \cite{ho2020denoising}, and Large Language Model (LLM) \cite{brown2020language,bommasani2021opportunities}.
AEs enable the modeling of data distribution by learning a compressed representation of the input data and reconstructing the input from the compressed representation. 
Among them, the Variational Autoencoder (VAE) can generate new samples similar to the input data by introducing hidden variables and variational inference.
GANs learn through the adversarial learning of the generator and the discriminator, where the generator tries to generate samples that are as similar as possible to the real data distribution, while the discriminator tries to differentiate between the generated samples and the real samples. The game between the two finally enables the generator to generate high-quality samples.
DMs achieve sample generation by learning the process of gradual perturbation of the data distribution and by reversing the process, which has excellent performance in terms of generation quality and diversity.
LLMs, such as the Generative Pre-trained Transformer (GPT) series \cite{achiam2023gpt}, learn the statistical patterns and semantic representations of language by pre-training on vast amounts of text data. This pre-training process involves exposing the models to diverse and extensive corpora, allowing them to capture the intricate nuances, contextual dependencies, and latent structures inherent in natural language.
In terms of the structure of this survey, we will elucidate the mathematical principles underlying these models to help readers gain a comprehensive understanding of the development and evolution of generative techniques.

\footnotetext{
\begin{minipage}{\textwidth}
Facial image is sourced from \url{https://zenodo.org/records/1188976}. 

Feature extraction module is adapted from is from \url{https://medium.com/clique-org/how-to-create-a-face-recognition-model-using-facenet-keras-fd65c0b092f1}. 

Dimensional emotion models can be found at \cite{inproceedings}.

Data augmentation schematic is from \url{https://www.baeldung.com/cs/ml-gan-data-augmentation}. 

Semi-supervised learning schematic is provided by \cite{latif2020multi}.

Cross-domain image is from \cite{Nasersharif2023}.

Schematics of each of the four generative models are taken from \url{https://www.compthree.com/blog/autoencoder/}, \url{https://www.javatpoint.com/generative-adversarial-network},
\url{https://towardsdatascience.com/diffusion-models-made-easy-8414298ce4da},
\url{https://engineeringprompts.substack.com/p/frameworks-to-build-llm-applications}
\end{minipage}%
}

Then, this survey will cover datasets commonly used in the field of generative technology for emotion recognition. 
For emotion recognition based on speech, facial images, and textual information, 
a variety of datasets will be discussed. 
Particularly,
FER2013 \cite{goodfellow2013challenges} provides a wide range of facial expressions collected via the Google search engine and labeled for seven emotional states. 
AFEW \cite{6200254}, derived from movies, serves as a dynamic, real-world dataset capturing both auditory and visual expressions. 
IEMOCAP \cite{busso2008iemocap} consists of audio-visual recordings of acted emotional states.
RAVDESS \cite{livingstone2018ryerson} includes both voice and video recordings of actors performing emotional expressions in a controlled environment. 
Additionally, CMU-MOSEI \cite{zadeh2018multimodal} integrates audio, video, and text annotations, making it one of the most versatile datasets available for studying emotion recognition based on multiple modalities. 
For emotion recognition based on physiological signals, several datasets will be introduced. 
For example,
DEAP \cite{koelstra2011deap} includes EEG and peripheral physiological signals of participants as they respond to music videos, providing valuable insights into emotional responses. 
DREAMER \cite{katsigiannis2017dreamer} is another significant dataset that provides both EEG and ECG signals while subjects watch emotional video clips. 
By exploring these diverse datasets, readers will gain a better understanding of the characteristics and challenges associated with different emotion recognition tasks.

\begin{figure}[]
  \centering
  \includegraphics[width=0.7\textwidth]{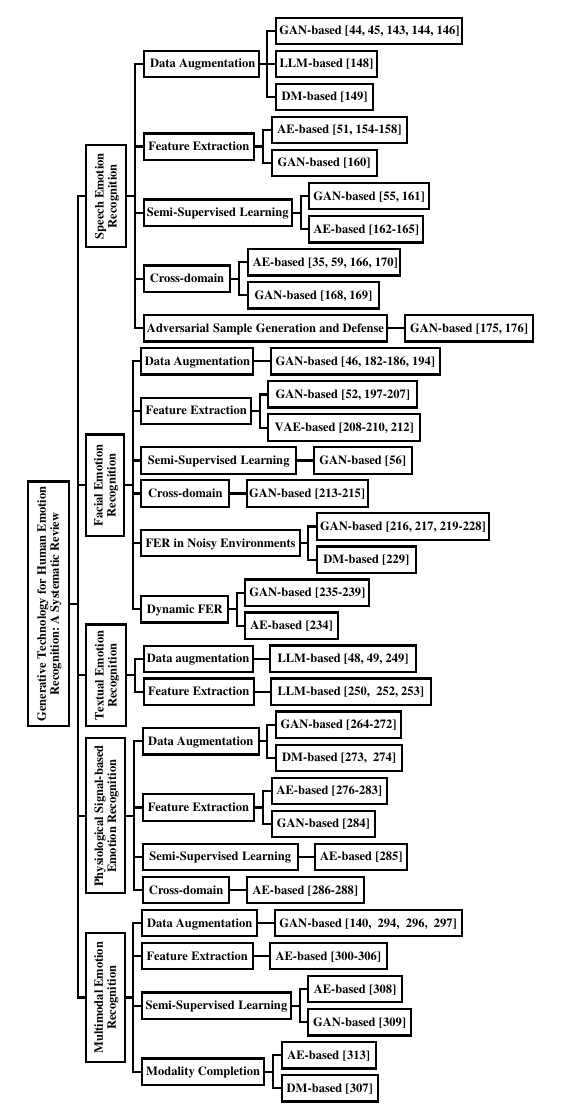}
  \caption{Taxonomy of This Survey.}
  \label{Overall structure}
\end{figure}

Based on the introduction of the principles of generative models and the datasets used, this review will show the research progress of generative models in the field of emotion recognition from various perspectives, including data augmentation, feature extraction, semi-supervised learning, cross-domain, and so on. 
The taxonomy is shown in Figure \ref{Overall structure}. 
Regarding data augmentation, it is an important way to improve the generalization performance of the model. 
However, traditional data augmentation methods, such as rotation and cropping, are difficult to portray the intrinsic attributes of emotional data \cite{maharana2022review}. 
Generative models open a new path for sentiment data augmentation. 
For example, in SER, researchers \cite{chatziagapi2019data,A2022} use GAN to generate samples similar to real speech emotion data, enhancing the model's ability to adapt to different speech conditions. 
In FER, researchers \cite{porcu2020evaluation,li2023semantic} integrate GAN and VAE to generate facial expression images under different pose, light, and occlusion conditions, improving the generalization ability of the model. 
In TER, researchers \cite{nedilko2023generative,koptyra2023clarin} utilize LLMs to generate text samples with different emotional characteristics, expanding the diversity of training data.
Feature extraction here refers to the use of generative models to learn effective feature representations from emotional data \cite{harshvardhan2020comprehensive}.
For instance, 
AEs and GANs can be used to learn compact representations from speech and facial expressions. 
These learned features can be used in downstream emotion recognition tasks to improve recognition performance \cite{Latif2017,Yang2020}.
Semi-supervised learning \cite{van2020survey,yang2022survey} is a paradigm for jointly training models using a large amount of unlabeled data and a small amount of labeled data, which is important for alleviating the scarcity of labeled data. 
Generative models provide new technical tools for semi-supervised emotion recognition. 
For example, some researchers try to apply GANs to semi-supervised learning by synthesizing unlabeled samples using a generator, then predicting pseudo-labels using a discriminator, and finally co-training classifiers with pseudo-labeled samples and real labeled samples \cite{Zhao2020,chen2019emotion}.
Cross-domain refers to the phenomenon where the performance of emotion recognition models significantly degrades in different domains \cite{redko2020survey,farahani2021brief}. This is usually caused by the differences in data distribution between the source domain and the target domain. By capturing the commonalities between different domains, generative models can achieve cross-domain emotion recognition and improve the adaptability of models in new domains \cite{Nasersharif2023,Xiao2020}.

Finally, this review will present the main findings and provide an outlook on future research directions based on the above systematic analysis. The main findings include: 
(i) 
Generative models are most widely used in FER compared to emotion recognition based on other modalities.
(ii) 
Generative models are primarily applied to emotion recognition in the form of data augmentation and feature extraction, although the specific working mechanisms may differ.
(iii) 
AEs and GANs are more widely used than other generative models.
Looking ahead, there are still many directions worth exploring, which include: (i) 
Investigating how to combine DMs and Transformer architecture for emotion recognition. This is expected to be a hotspot for future research, as these have already achieved significant success in a number of domains and may provide new opportunities to improve emotion recognition performance \cite{peebles2023scalable,bao2023all}.
(ii) Combining techniques such as reinforcement learning \cite{arulkumaran2017deep} and federated learning \cite{li2020review} with generative models to better model the complex dependencies and temporal evolution of emotions. 
This highly promising direction has the potential to capture the intricate dynamics and context-dependent nature of emotional expressions, leading to more accurate and nuanced emotion recognition systems.
(iii) Further expanding the combination of generative technology in emotion recognition, particularly in virtual reality (VR) and augmented reality (AR) applications \cite{marin2020emotion,papoutsi2021virtual}, to enhance user experiences. 
(iv) 
Based on the generative models discussed in this paper, it is possible to directly generate
natural, realistic content with rich emotional expressions, such as emotionally charged speech, images, text, and videos \cite{wang2023emotional,lian2023gpt}. 
To sum up,
this relatively unexplored area presents a wealth of opportunities for innovation and creative applications of generative models, with the potential to improve how emotions are detected, interpreted, and responded to in immersive environments.
Through the above work, this survey aims to deepen the understanding of generative technology in the context of emotion recognition and provide inspiration and guidance for future research. 
Overall, the main contributions of this survey include:

(1) To the best of our knowledge, this paper is the first work to systematically review the research progress of generative technology in the field of emotion recognition, filling a gap in the existing literature.

(2) By analyzing more than 320 research papers, this paper gives a taxonomy of generative models for emotion recognition from different perspectives, including data augmentation, feature extraction, semi-supervised learning, cross-damain, etc. 
This in-depth analysis enables readers to gain a thorough understanding of the diverse applications and methodologies of generative technology in emotion recognition.

(3) 
We summarize the benchmark datasets used according to the different modalities, highlighting key characteristics such as sample size, number of subjects, and emotion categories. Furthermore, we present the performance of different generative models on emotion recognition based on these datasets.

(4) Finally, we discuss some findings of generative models for emotion recognition and further point out potential future research directions.

The rest of the paper is structured as follows: Section \ref{diff} introduces the difference between this review and other existing reviews, 
Section \ref{gm} gives the mathematical principles of different generative models, 
Section \ref{dataset} describes the datasets used, 
Sections \ref{ser} - \ref{multi er} present a series of work on SER, FER, TER, emotion recognition based on physiological signals, and MER based on generative models, respectively,  
Section \ref{discussion} gives the main findings and future outlook, and Section \ref{conclusion} gives the conclusion. A list of abbreviations is given in Table \ref{Main acronyms}.

\begin{table}[H]
\centering
\caption{Main Acronyms.}
\label{Main acronyms}
\resizebox{\textwidth}{!}{%
\begin{tabular}{|l|l|l|l|}
\hline
Acronym & Full Form                             & Acronym & Full Form                               \\ \hline
SER     & Speech Emotion Recognition            & 
FER     & Facial Emotion Recognition
                       \\ \hline
TER     & Textual Emotion Recognition  &     
MER     & Multimodal Emotion Recognition
                 \\ \hline
AE      & Autoencoder                           & DAE     & Denoising Autoencoder   
\\ \hline
AAE     & Adversarial Autoencoder  & 
VAE     & Variational Autoencoder 
\\ \hline
SAE     & Stacked Autoencoder  & 
CAE     & Convolutional Autoencoder               \\ \hline
GAN     & Generative Adversarial Network  &
DCGAN   & Deep Convolutional GAN                  \\ \hline
CGAN    & Conditonal GAN  & 
ACGAN   & Auxiliary Classifier GAN                \\ \hline
DANN   & Domain Adversarial Neural   Network  &
GCNN & Graph Convolutional Neural   Network \\ \hline
DM      & Diffusion Model         & 
LLM     & Large Language Model    \\ \hline
DDPM & Denoising Diffusion Probabilistic Model &
ViT     & Vision Transformer \\ \hline
GPT     & Generative Pre-trained   Transformer  &
T5      & Text-to-Text Transfer   Transformer \\ \hline
MLP     & Multilayer Perceptron                 &
CNN     & Convolutional Neural Networks         \\ \hline
RNN     & Recurrent Neural Network              & 
LSTM    & Long Short-Term Memory          \\ \hline
MFCC   & Mel-Frequency Cepstral   Coefficient &
AU     & Action Unit 
\\ \hline
EEG     & Electroencephalogram                   &
ECG     & Electrocardiogram                       \\ \hline
EMG     & Electromyogram                         &
HRV     & Heart Rate Variability                  \\ \hline
EDA     & Electrodermal Activity                 & fMRI    & functional Magnetic Resonance Imaging \\ \hline
AUC     & Area Under Curve             & CCC     & Consistency Correlation Coefficient     \\ \hline
ACC     & Accuracy                              & 
UAR     & Unweighted Average Recall                            \\ \hline
\end{tabular}%
}
\end{table}

\section{The Difference}
\label{diff}
\quad
Existing reviews of generative technology mainly focus on GAN-based generation of single modalities such as images, speech, and text.
For example, 
Kammoun et al. \cite{kammoun2022generative} provide an overview of advances in GANs for face generation, focusing on their applications, architectures, training modes, and evaluation metrics. 
Wali et al. \cite{wali2022generative} categorize speech GANs according to application areas: speech synthesis, speech enhancement and conversion, and data augmentation in speech recognition and emotional speech recognition systems, and summarize commonly used datasets and evaluation metrics in speech GANs.
In particular, the review by Hajarolasvadi et al. \cite{hajarolasvadi2020generative} covers various aspects of GAN-based human emotion synthesis, including facial expression synthesis and speech emotion synthesis.

Emotion recognition, as a much-anticipated area in affective computing, 
has been reviewed by researchers in a number of aspects.
For example, 
Deng and Ren \cite{deng2021survey} provide a comprehensive overview of TER research, focusing on deep learning approaches, which categorizes TER methods based on different stages of implementation, such as word embeddings, architectures, and training levels.
Zhao et al. \cite{zhao2021survey} present the history of FER, emotion representation models, well-known datasets, and applications of FER in fields such as transportation, healthcare, and business.
Khare et al. \cite{khare2023emotion} present emotion models based on different modalities, stimuli used for emotion elicitation, and existing automatic emotion recognition systems.
Cîrneanu et al. \cite{cirneanu2023new} discuss different neural network architectures used in FER, including Multilayer Perceptrons (MLPs), Convolutional Neural Networks (CNNs), Recurrent Neural Networks (RNNs), and emphasize the key elements and performance of each architecture.
Younis et al. \cite{younis2024machine} provide an overview of the advancements in emotion recognition from the broad perspective of machine learning, but lack a detailed analysis of the progress made using generative models.

Through the above analysis, it can be seen that existing works conduct separate reviews of generative technology and emotion recognition. 
However, due to the superior characteristics of generative models, they begin to be widely applied in the field of emotion recognition, such as data augmentation \cite{nanthini2023survey}, feature extraction \cite{eigenschink2023deep}, etc.  
Nonetheless, so far, there is no work that systematically reviews generative models for emotion recognition.
This paper aims to fill this gap by comprehensively summarizing the latest progress and applications of generative models in the field of emotion recognition, providing references and insights for future research.
Table \ref{Differences between ours and other reviews} shows how our work differs from other reviews.

\begin{table}[H]
\centering
\caption{Differences Between Ours and Previous Reviews.}
\label{Differences between ours and other reviews}
\resizebox{\textwidth}{!}{%
\begin{tabular}{|c|c|c|c|}
\hline
Reference &
  \multicolumn{1}{c|}{\begin{tabular}[c]{@{}c@{}}{Focus on}\\ generative models\end{tabular}} &
  \multicolumn{1}{c|}{\begin{tabular}[c]{@{}c@{}}Related to\\ emotion recognition\end{tabular}} &
  \multicolumn{1}{c|}{\begin{tabular}[c]{@{}c@{}}Focus on\\ multiple modalities\end{tabular}} \\ \hline
Ours                                                 & \checkmark & \checkmark & \checkmark \\ \hline
Kammoun et al.   \cite{kammoun2022generative}             & \checkmark &            &            \\ \hline
Wali et al. \cite{wali2022generative}                     & \checkmark & \checkmark &            \\ \hline
Hajarolasvadi et al.   \cite{hajarolasvadi2020generative} & \checkmark &            &            \\ \hline
Deng and Ren \cite{deng2021survey}                        &            & \checkmark &            \\ \hline
Zhao et al. \cite{zhao2021survey}                         &            & \checkmark &            \\ \hline
Khare et al. \cite{khare2023emotion}                      &            & \checkmark & \checkmark \\ \hline
Cîrneanu et al. \cite{cirneanu2023new}                    & \checkmark & \checkmark &            \\ \hline
Younis et al. \cite{younis2024machine}                    &            & \checkmark &
\checkmark   \\ \hline
\end{tabular}%
}
\end{table}

\section{Generative Model}
\label{gm}
\quad 
Generative models are a powerful class of machine learning models that aim to learn the underlying probability distribution of a given dataset. 
By capturing the intricate patterns and structures within the data, these models can generate new samples that closely resemble the original data distribution. The ability to create realistic and diverse examples has led to a wide range of applications across various fields \cite{oussidi2018deep,cao2024survey}. 
Generative models can be achieved in different approaches to model the probability distribution of data.
Common generative modeling methods include AEs, GANs, DMs, and LLMs.
These methods have their own unique characteristics in modeling complex data distributions and enhancing model generation capabilities \cite{bond2021deep}.
The following describes each of these models in detail. 

\subsection{Autoencoder (AE)} 
\quad
AE \cite{rumelhart1986learning} is an unsupervised generative model that learns the compressed representation of data and reconstructs the input data. 
It consists of an encoder that maps the original input data to a low-dimensional latent space and a decoder that maps the latent representation back to the original data space. Typical variants of AE include Adversarial Autoencoder (AAE) and Variational Autoencoder (VAE). Unlike traditional AE, AAE \cite{makhzani2015adversarial} introduces a discriminator network that evaluates the authenticity of reconstructed data by comparing the encoded latent representation with randomly sampled vectors from a prior distribution. 

VAE \cite{kingma2013auto} is a generative model that combines the ideas of AE and variational inference. 
It works as follows: 
First, the input data is passed through an encoder,
which maps the input data to the mean and variance parameters in the latent space.
Then, a hidden variable is sampled from the latent space, usually using a normal distribution.
Next, the sampled hidden variable is passed through a decoder, which maps the hidden variable back to the space of the original input data to generate a reconstructed sample. 
To constrain the latent representation, VAE uses the Kullback-Leibler (KL) divergence as a regularization term to make the learned latent representation closer to a prior distribution. 
Therefore, the VAE's loss function consists of a reconstruction error term and a KL divergence term, given by:
\begin{equation}
L = -\mathbb{E}_{q(z|x)}[\log p(x|z)] + \textup{KL}(q(z|x) || p(z))\label{eq:loss_vae}
\end{equation}
where $\mathbb{E}_{q(z|x)}$ indicates the calculation of the expectation over the posterior distribution $q(z|x)$ of the latent variable $z$ generated by the encoder.
$\log p(x|z)$ refers to the reconstruction error term, which indicates the log-likelihood of generating data $x$ given the latent variable $z$.
$\textup{KL}(q(z|x) || p(z))$ represents the KL divergence term, which is used to evaluate the dissimilarity between the latent variable distribution generated by the encoder $q(z|x)$ and the prior distribution $p(z)$.

\subsection{Generative Adversarial Network (GAN)}
\quad 
GAN \cite{goodfellow2014generative} is a generative model consisting of two components: a generator and a discriminator. 
They learn from each other in an adversarial process, where the generator aims to generate more realistic samples, and the discriminator aims to distinguish between the generated samples and real samples. 
In addition to the traditional GAN model, there are several variants of GANs, such as Conditional GANs (CGANs) \cite{mirza2014conditional}, Deep Convolutional GANs (DCGANs) \cite{radford2015unsupervised}, Wasserstein GANs \cite{arjovsky2017wasserstein}, and CycleGANs \cite{zhu2017unpaired}.

To be specific, the training process of GAN involves an iterative optimization procedure. 
The generator takes a random noise vector as input and generates a sample.
The discriminator then takes both real and generated samples as input, and outputs a probability that represents the likelihood of the input being a real sample. 
The parameters of the generator and discriminator are updated separately to improve their performance.
The generator's loss function is designed to maximize the probability of the discriminator classifying the generated samples as real, which 
can be formulated as:
\begin{equation}
L_G = -\mathbb{E}_{z \sim p_z(z)}[\log(D(G(z)))]
\end{equation}
where $G$ represents the generator, $D$ represents the discriminator, $z$ is a random noise vector sampled from a prior distribution $p_z(z)$ (e.g., a Gaussian distribution), and $\mathbb{E}$ denotes the expected value.
On the other hand, the discriminator's loss function is designed to maximize the probability of correctly classifying both real and generated samples, which can be expressed as:
\begin{equation}
L_D = -\mathbb{E}_{x \sim p_{data}(x)}[\log(D(x))] - \mathbb{E}_{z \sim p_z(z)}[\log(1 - D(G(z)))]
\end{equation}
where $x$ represents real samples from the training data distribution $p_{data}(x)$.
This dynamic equilibrium results in the generator being able to generate samples that resemble real data closely, making GAN capable of generation tasks.

\subsection{Diffusion Model (DM)}
\quad 
DMs \cite{ho2020denoising} are a class of generative models which have gained significant attention in recent years due to their ability to generate high-quality and diverse samples. Unlike other generative models that directly learn the data distribution, DMs learn to transform a simple noise distribution into the desired data distribution through a gradual diffusion process.
The core idea behind DMs is to define a forward diffusion process that gradually adds noise to the data, and then learn a reverse diffusion process that removes the noise step by step, eventually generating clean samples. The diffusion process is typically defined as a Markov chain, where each step corresponds to a small amount of Gaussian noise being added to the data.

The forward diffusion process can be described by a sequence of latent variables $z_0$, \dots, $z_T$, 
where $z_0$ represents the clean data and $z_T$ represents the fully noised data. 
The transition from $z_t$ to $z_{t+1}$ is modeled by a Gaussian transition kernel $q(z_{t+1}|z_t)$, which adds Gaussian noise with a predetermined variance schedule.
The reverse diffusion process, also known as the denoising process, is learned by training a neural network to predict the noise that was added in each step of the forward diffusion process. Given a noisy sample $z_t$ at time step $t$, the goal is to learn a function $f_\theta(z_t, t)$ that predicts the noise $\epsilon_t$, such that the clean data can be recovered by subtracting the predicted noise from the noisy sample:
\begin{equation}
z_{t-1} = z_t - \sqrt{\beta_t} \cdot \epsilon_t
\end{equation}
where $\beta_t$ is a time-dependent noise scaling factor.
The training objective of DMs is to minimize the weighted sum of the squared error between the predicted noise and the actual noise added in each step of the forward diffusion process. 
This is typically achieved using a variant of the variational lower bound, such as the evidence lower bound (ELBO) or the simplified loss proposed in \cite{ho2020denoising}.

During inference, the generation process starts with a sample from the noise distribution $z_T$ and iteratively applies the learned denoising function $f_\theta(z_t, t)$ to remove the noise and obtain increasingly cleaner samples. The generated sample at each step is obtained by:
\begin{equation}
z_{t-1} = z_t - \sqrt{\beta_t} \cdot f_\theta(z_t, t)
\end{equation}
By repeating this process for a sufficient number of steps, DMs can generate high-quality samples that closely resemble the training data.

\subsection{Large Language Model (LLM)}
\quad 
LLMs are a type of deep learning model based on the principles of generative modeling, aiming to learn the probability distribution of words in text data to generate coherent and natural text sequences \cite{brown2020language,bommasani2021opportunities}.
The basic principle involves pre-training on large-scale text data to capture semantic relationships and syntactic structures between words, followed by fine-tuning on specific tasks to adapt to particular application scenarios \cite{brown2020language}. 
They typically employ neural network architectures like the Transformer \cite{vaswani2017attention} model. 
During pre-training, these models utilize unsupervised learning on unlabeled text data to learn rich language representations \cite{radford2018improving}.
In the fine-tuning stage, LLMs improve their performance on specific tasks through supervised learning \cite{raffel2020exploring}.
LLMs, such as T5 (Text-to-Text Transfer Transformer) \cite{raffel2020exploring}, XLnet \cite{yang2019xlnet}, LLaMA \cite{touvron2023llama}, and GPT (Generative Pre-trained Transformer) \cite{achiam2023gpt}, demonstrate significant potential in various natural language processing tasks, leading to advancements in text generation \cite{zhao2023survey}.

\vspace{1em}
\quad 
Through the above discussion, we explore the mathematical principles
behind several representative generative models, including AE, GAN, DM, and LLM. These models exhibit unique characteristics in learning complex data distributions and generating realistic samples, providing powerful tools for applications in various domains, including emotion recognition \cite{wali2022generative,cao2024survey}.

\section{Databases}
\label{dataset}
\quad
This section focuses on the datasets commonly used in generative technology for emotion recognition. These datasets are crucial in this field because the performance of related algorithms largely depends on the quality and richness of the data. 
Table \ref{databases} lists these datasets, which capture and represent emotional information in various ways, enabling researchers to develop and evaluate generative models that can effectively understand and generate emotional content. Specifically, as shown in Table \ref{databases}, these datasets can be divided into two main categories: spontaneous datasets and in-the-wild datasets \cite{guerdelli2022macro}. 
Spontaneous datasets include expressions simulated by participants,  
where emotions are expressed naturally despite participants' awareness that they are being monitored.
The acquisition environment for these datasets is typically a controlled laboratory setting. 
On the other hand, in-the-wild datasets capture emotions in real-world scenarios without the need for controlled or processed collection, and participants are filmed in their natural environments. 
In addition, ``Modalities" in Table \ref{databases} refers to the modalities contained in the dataset, such as visual or physiological data. 
``Samples" indicates the type and number of samples in the dataset. 
``Subjects" refers to the number of individuals included in the dataset.
``Categories" shows the categories for emotion recognition, indicating the range of emotions that the dataset covers.
A slash indicates that the specific information is not explicitly mentioned in the dataset description. 
For example, the IAPS dataset \cite{lang1997international} contains data in the form of visual modality, including over 9,000 images, and the emotion category information includes valence, arousal, and dominance. 
In the following sections, we will show the performance of different generative models for emotion recognition based on the datasets listed in Table \ref{databases}.

\begin{landscape}
\begin{longtable}{|>{\raggedright\arraybackslash}p{2cm}|>{\raggedright\arraybackslash}p{1cm}|>{\raggedright\arraybackslash}p{3.5cm}|>{\raggedright\arraybackslash}p{2.5cm}|>{\raggedright\arraybackslash}p{3cm}|>{\raggedright\arraybackslash}p{1.5cm}|>{\raggedright\arraybackslash}p{9cm}|}
\caption{Common Databases for Emotion Recognition with Generative Models.}
\label{databases}\\
\hline
Database &
  Year &
  Modalities &
  \begin{tabular}[c]{@{}l@{}}Spontaneous/\\      in-the-wild\end{tabular} &
  Samples &
  Subjects &
  Categories \\ \hline
\endfirsthead
\multicolumn{7}{c}%
{{\bfseries Table \thetable\ continued from previous page}} \\
\hline
Database &
  Year &
  Modalities &
  \begin{tabular}[c]{@{}l@{}}Spontaneous/\\      In The Wild\end{tabular} &
  Samples &
  Subjects &
  Category \\ \hline
\endhead
IAPS \cite{lang1997international} &
  1997 &
  Visual &
  In the Wild &
  More than 9000 images &
  / &
  Valence, Arousal, and Dominance \\ \hline
TFD \cite{susskind2010toronto} &
  2010 &
  Visual &
  In the Wild &
  4178 images &
  / &
  Happiness,  Sadness, Surprise,   Fear, Anger, Disgust, Neutral \\ \hline
SFEW \cite{dhall2011static} &
  2011 &
  Visual &
  In the Wild &
  1739 images &
  330 &
  Happiness,  Sadness, Surprise,   Fear, Anger, Disgust, Neutral \\ \hline
FER2013 \cite{goodfellow2013challenges} &
  2013 &
  Visual &
  In the Wild &
  35887 images &
  / &
  Happiness,  Sadness, Surprise,   Fear, Anger, Disgust, Neutral \\ \hline
AffectNet   \cite{mollahosseini2017affectnet} &
  2017 &
  Visual &
  In the Wild &
  450000 images &
  / &
  Happiness,  Sadness, Surprise,   Fear, Anger, Disgust, Neutral, Contempt \\ \hline
RAF-DB \cite{shan2018reliable} &
  2017 &
  Visual &
  In the Wild &
  29672 images &
  / &
  Happiness,  Sadness, Surprise,   Fear, Anger, Disgust, Neutral, Contempt \\ \hline
JAFFE \cite{lyons1998coding} &
  1998 &
  Visual &
  Spontaneous &
  219 images &
  10 &
 Happiness,  Sadness, Surprise,   Fear, Anger, Disgust, Neutral \\ \hline
BU-3DFE \cite{yin20063d} &
  2006 &
  Visual &
  Spontaneous &
  2500 images &
  100 &
  Happiness,  Sadness, Surprise,   Fear, Anger, Disgust, Neutral \\ \hline
TFEID\cite{chen2007taiwanese} &
  2007 &
  Visual &
  Spontaneous &
  7200 images &
  40 &
  Happiness,  Sadness, Surprise,   Fear, Anger, Disgust, Neutral, Contempt \\ \hline
Multi-PIE \cite{gross2010multi} &
  2008 &
  Visual &
  Spontaneous &
  750000 images &
  337 &
  Happiness, Surprise, Sadness, Anger, Fear, Disgust \\ \hline
BU-4DFE \cite{zhang2013high} &
  2008 &
  Visual &
  Spontaneous &
  606 sequences &
  101 &
  Happiness, Surprise, Sadness, Anger, Fear, Disgust \\ \hline
FAU Aibo \cite{batliner2008releasing} &
  2008 &
  Audio &
  Spontaneous &
  9.2 hours of speech &
  31 &
  Joyful, Surprised, Emphatic, Helpless, Touchy, Rest \\ \hline
CHB-MIT \cite{shoeb2009application} &
  2009 &
  EEG &
  Spontaneous &
  / &
  23 &
  Arousal and Valence \\ \hline
USTC-NVIE\cite{wang2010natural}&
  2010 &
  Visual &
  Spontaneous &
  236 images &
  215 &
  Happiness,  Sadness, Surprise,   Fear, Anger, Disgust, Neutral \\ \hline
CK+ \cite{lucey2010extended} &
  2010 &
  Visual &
  Spontaneous &
  593 images &
  123 &
  Happiness,  Sadness, Surprise,   Fear, Anger, Disgust, Neutral, Contempt \\ \hline
FACES \cite{ebner2010faces} &
  2010 &
  Visual &
  Spontaneous &
  1026 videos &
  171 &
  Happiness, Surprise, Sadness, Anger, Fear, Disgust \\ \hline
Oulu-CASIA\cite{zhao2011facial} &
  2011 &
  Visual &
  Spontaneous &
  480 sequences &
  80 &
  Happiness, Surprise, Sadness, Anger, Fear, Disgust \\ \hline
SMIC \cite{li2013spontaneous} &
  2013 &
  Visual &
  Spontaneous &
  164 sequences &
  16 &
  Positive, Negative, Surprise \\ \hline
SEED \cite{duan2013differential} &
  2013 &
  EEG &
  Spontaneous &
  / &
  15 &
  Positive, Neutral, Negative \\ \hline
CFEE \cite{du2014compound} &
  2014 &
  Visual &
  Spontaneous &
  229 images &
  230 &
  Happiness, Surprise, Sadness, Anger, Fear, Disgust \\ \hline
CASME II \cite{yan2014casme} &
  2014 &
  Visual &
  Spontaneous &
  255 sequences &
  35 &
  Happiness, Surprise, Disgust, Repression, and Others \\ \hline
ECG 200 \cite{olszewski2001generalized} &
  2001 &
  ECG &
  Spontaneous &
  200 samples &
  / &
  Happiness, Surprise, Sadness, Anger, Fear, Disgust \\ \hline
SAMM \cite{davison2016samm} &
  2016 &
  Visual &
  Spontaneous &
  159 sequences &
  32 &
  Happiness,  Sadness, Surprise,   Fear, Anger, Disgust, Contempt \\ \hline
ADFES-BIV \cite{wingenbach2016validation} &
  2016 &
  Visual &
  Spontaneous &
  320 videos &
  12 &
  \begin{tabular}[c]{@{}l@{}}Anger, Contempt, Disgust, Embarrassment, Fear,  \\ Pride, Joy, Neutral, Sad, Surprise \end{tabular} \\ \hline
BP4D+ \cite{zhang2016multimodal} &
  2016 &
  Visual &
  Spontaneous &
  about 1.4 million frames &
  140 &
  12 Different Categories and 5-point Likert-type Scales \\ \hline
KDEF \cite{calvo2018human} &
  2018 &
  Visual &
  Spontaneous &
  490 images &
  272 &
  Happiness,  Sadness, Surprise,   Fear, Anger, Disgust, Neutral \\ \hline
AudiofMRI Parallel Set   \cite{pereira2018toward} &
  2018 &
  fMRI &
  Spontaneous &
  / &
  18 &
  Sad, Happy, Excited, Surprise, Neutral, Angry, Distress, Frustrated \\ \hline
F2ED \cite{wang2019fine} &
  2019 &
  Visual &
  Spontaneous &
  219719 images &
  119 &
  54 Different Facial Expression Classes \\ \hline
Affwild \cite{kollias2019deep} &
  2019 &
  Visual &
  In the Wild &
  298 videos &
  / &
  Arousal and Valence \\ \hline
AFEW \cite{6200254} &
  2012 &
  Visual + Audio &
  In the Wild &
  1426 sequences &
  330 &
  Happiness,  Sadness, Surprise,   Fear, Anger, Disgust, Neutral \\ \hline
EMO-DB \cite{burkhardt2005database} &
  2005 &
  Audio + Text &
  Spontaneous &
  500 utterances &
  10 &
  Neutral, Anger, Fear, Joy, Sadness, Disgust, Boredom \\ \hline
\begin{tabular}[c]{@{}l@{}}eNTERF-\\ ACE’05 \cite{martin2006enterface}\end{tabular} &
  2006 &
  Visual + Audio &
  Spontaneous &
  1166 sequences &
  42 &
  Happiness,  Sadness, Surprise,   Fear, Anger, Disgust, Neutral \\ \hline
SAVEE\cite{haq2008audio} &
  2008 &
  Visual + Audio &
  Spontaneous &
  480 utterances &
  4 &
  Happiness,  Sadness, Surprise,   Fear, Anger, Disgust, Neutral \\ \hline
VAM \cite{grimm2008vera} &
  2008 &
  Audio + Text &
  Spontaneous &
  12 hours of data &
  47 &
  Happiness, Surprise, Sadness, Anger, Fear, Disgust \\ \hline
MMI \cite{valstar2010induced} &
  2010 &
  Visual + Audio &
  Spontaneous &
  2900 videos &
  75 &
  Happiness, Surprise, Sadness, Anger, Fear, Disgust \\ \hline
RECOLA \cite{ringeval2013introducing} &
  2013 &
  Visual + Audio &
  Spontaneous &
  1308 utterances &
  23 &
  Valence and Arousal \\ \hline
CREMA-D \cite{cao2014crema} &
  2014 &
  Visual + Audio &
  Spontaneous &
  7442 utterances &
  91 &
  Happiness, Surprise, Sadness, Anger, Fear, Disgust \\ \hline
EMOVO \cite{costantini2014emovo} &
  2014 &
  Audio + Text &
  Spontaneous &
  84 sentences &
  6 &
  Happiness, Surprise, Sadness, Anger, Fear, Disgust \\ \hline
JTES \cite{takeishi2016construction} &
  2016 &
  Audio + Text &
  Spontaneous &
  20,000 speech samples &
  100 &
  Neutral, Angry, Joy, Sad \\ \hline
MSP-IMPROV \cite{7374697} &
  2017 &
  Visual + Audio &
  Spontaneous &
  8348 samples &
  12 &
  Sadness, Happiness, Anger, Neutrality \\ \hline
DREAMER \cite{katsigiannis2017dreamer} &
  2017 &
  EEG + ECG &
  Spontaneous &
  / &
  25 &
  Arousal and Valence \\ \hline
RAVDESS \cite{livingstone2018ryerson} &
  2018 &
  Visual + Audio &
  Spontaneous &
  7356 viseos &
  24 &
  Happiness,  Sadness, Surprise,   Fear, Anger, Disgust, Neutral, Contempt \\ \hline
URDU \cite{latif2018cross} &
  2018 &
  Audio + Text &
  Spontaneous &
  400 utterances &
  38 &
  Angry, Happy, Sad, Neutral \\ \hline
Emoti-W \cite{dhall2017individual} &
  2017 &
  Visual + Audio + Text &
  In the Wild &
  More than 100 movies &
  / &
  Happiness,  Sadness, Surprise,   Fear, Anger, Disgust, Neutral \\ \hline
MELD \cite{poria2018meld} &
  2018 &
  Visual + Audio + Text &
  In the Wild &
  more than 1400 dialogues &
  304 &
  Happiness,  Sadness, Surprise,   Fear, Anger, Disgust, Neutral \\ \hline
CMU-MOSEI\cite{zadeh2018multimodal} &
  2018 &
  Visual + Audio + Text &
  In the Wild &
  videos and 23453 sentences &
  1000 &
  Happiness, Surprise, Sadness, Anger, Fear, Disgust \\ \hline
OMG \cite{barros2018omg} &
  2018 &
  Visual + Audio + Text &
  In the Wild &
  7371 utterances &
  / &
  Happiness,  Sadness, Surprise,   Fear, Anger, Disgust, Neutral \\ \hline
IEMOCAP \cite{busso2008iemocap} &
  2008 &
  Visual + Audio + Text &
  Spontaneous &
  10039 samples &
  10 &
  Arousal and Valence \\ \hline
DEAP \cite{koelstra2011deap} &
  2011 &
  \begin{tabular}[c]{@{}l@{}}EEG + EOG +\\ EMG + GSR\end{tabular} &
  Spontaneous &
  / &
  32 &
  Arousal and Valence \\ \hline
UlmTSST \cite{stappen2021muse} &
  2021 &
  \begin{tabular}[c]{@{}l@{}}EDA + ECG + \\ RESP + BPM\end{tabular} &
  Spontaneous &
  / &
  69 &
  Arousal and Valence \\ \hline
\end{longtable}
\end{landscape}

\section{Speech Emotion Recognition (SER)}
\label{ser}

\quad 
SER is a crucial branch of emotion recognition that aims to identify and understand the emotional state of a speaker by analyzing the features in speech signals \cite{akccay2020speech}. Speech signals contain rich emotional information, such as intonation, volume, and speaking rate, which can reflect the speaker's emotional changes \cite{khalil2019speech}. 
In recent years, generative models have contributed to the development of SER in multiple aspects. 
In the following sections, we will delve into the specific applications of generative models in SER. 
Section \ref{5.1}
will focus on discussing speech data augmentation based on generative models. 
Section \ref{5.2} will explore the application of generative models for speech feature extraction. 
Section \ref{5.3} will describe the application of generative models in conjunction with semi-supervised learning, 
Section \ref{cdser} will present cross-domain SER based on generative models,
and Section \ref{5.4} 
will provide an introduction to the use of generative models for adversarial sample generation and defense.

\subsection{Speech Data Augmentation}
\label{5.1}
\quad 
Data augmentation is a crucial topic in emotion recognition, as it addresses the problem of limited training data \cite{maharana2022review,mumuni2022data}. 
In many real-world scenarios, acquiring and annotating large amounts of emotional data can be time-consuming, expensive, and labor-intensive. 
Moreover, emotional data may also suffer from the problem of class imbalance, where the number of labeled samples for certain emotional categories is relatively small, resulting in poor performance of the model on these categories \cite{krawczyk2016learning}.
The scarcity of labeled data described above
can hinder the development of robust and generalizable emotion recognition models.
Data augmentation techniques aim to mitigate this issue by artificially increasing the size and diversity of the training dataset, thereby improving the model's ability to learn and generalize from the available data. 
Traditionally, researchers have proposed various strategies for data augmentation in SER \cite{ko2015audio}. These strategies include simple techniques such as noise addition, pitch shifting, time stretching, and speed perturbation. 
These methods aim to introduce variations in the speech signal while preserving the emotional content. However, these traditional approaches often rely on handcrafted rules and may not capture the complex and nuanced patterns of emotional speech.

\begin{figure}[t]
  \centering
  \includegraphics[width=1\textwidth]{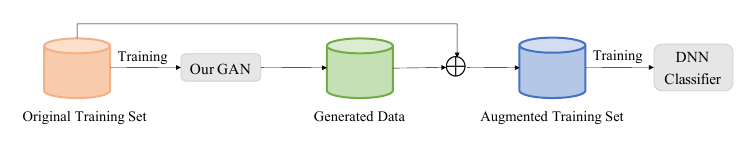}
  \caption{The pipeline of data augmentation from \cite{ma2022data}: First, a generative model is employed to generate new data, then the original training set is combined with the generated data, creating an augmented training set for the emotion classification task.
  }
  \label{dataaugmentation}
\end{figure}

In recent years, generative models have emerged as a promising approach for data augmentation in SER \cite{chatziagapi2019data,Heracleous2022}. 
By leveraging the power of generative models, researchers can create realistic and diverse emotional speech samples, effectively expanding the training dataset.
Figure \ref{dataaugmentation} illustrates a general architecture. 
Table \ref{da ser} presents different generative methods used for data augmentation in SER and their performance, which is typically evaluated using metrics such as Accuracy (ACC) and Unweighted Average Recall (UAR), where 
ACC provides an overall measure of the model's correctness, indicating its general performance, while UAR, 
calculates the average recall score across all classes, making it particularly useful when dealing with imbalanced datasets.

To be specific, generative models can effectively augment speech data by directly generating synthetic samples.
For example, 
Chatziagapi et al. \cite{chatziagapi2019data} adapt and improve the Balancing GAN architecture to generate realistic emotional data for minority classes.
Heracleous et al. \cite{Heracleous2022} propose a data augmentation framework that utilizes CycleGAN to transfer natural speech to emotional speech and employs a Vision Transformer (ViT) \cite{dosovitskiy2020image} as the classifier for SER.
Wang et al. \cite{Wang2022} 
proposes a GAN-based augmentation strategy with triplet loss to increase and stabilize the augmentation performance for the SER task.

In addition to directly performing augmentation in the data space for SER, some researchers attempt to conduct data augmentation in the feature space. 
This approach aims to manipulate and transform the features extracted from the raw speech data, generating new feature representations to increase the diversity and robustness of the training data.
For example,
in \cite{A2022}, a CycleGAN model is employed for data augmentation, and two feature selection networks, Fisher and Linear Discriminant Analysis, are used to enhance SER performance. 
Yi and Mak \cite{yi2020improving} combine GANs and AEs to design an adversarial data augmentation network to generate emotional feature vectors that share a common latent representation.
Latif et al. \cite{latif2020augmenting} utilize a data augmentation technique called mixup \cite{Zhang2017} to enhance GAN's ability in representation learning and synthesis of feature vectors.
In another study, Sahu et al. \cite{sahu2018enhancing} employ GANs to learn the distribution of low-dimensional representations of high-dimensional feature vectors and CGAN to learn the distribution of high-dimensional feature vectors conditioned on labels.

Particularly, it is worth noting that apart from
directly performing data augmentation on the speech modality, another approach is to transfer information from other modalities to the speech modality for data augmentation, thereby assisting in improving the performance of SER. This cross-modality information transfer-based data augmentation method has received increasing attention from researchers in recent years.
For instance,
He et al. \cite{he2020image2audio} 
employ GAN and VAE to convert facial images into spectrograms to increase the number of sample data, thereby enhancing SER.
Ma et al. \cite{Ma2023} use a LLM,  GPT-4, to generate emotional text, then use Azure emotional TTS to synthesize emotional speech.
Malik et al. \cite{Malik2023} utilize an improved DM model to generate synthesized emotional data. 
They use text embeddings represented by bidirectional encoders from transformers (BERT) \cite{devlin2018bert} to adjust the DM model to generate high-quality synthesized emotional samples in the voices of different speakers.

\begin{table}[]
\centering
\caption{Literature on Generative Models for Data Augmentation in SER.}
\label{da ser}
\resizebox{\columnwidth}{!}{%
\begin{tabular}{|l|l|l|l|l|l|}
\hline
Reference                                     & Year & Augmentation Domain & Model              & Dataset              & Performance (\%)       \\ \hline
Chatziagapi et al. \cite{chatziagapi2019data} & 2019 & Sample Space        & GAN                & IEMOCAP              & UAR: 53.6              \\ \hline
Heracleous et al. \cite{Heracleous2022}       & 2022 & Sample Space        & CycleGAN           & RAVDESS/JTES         & ACC: 77/66.1           \\ \hline
Wang et al. \cite{Wang2022}                   & 2022 & Sample Space        & GAN                & IEMOCAP              & UAR: 58.19             \\ \hline
Shilandari et al. \cite{A2022}                & 2022 & Feature Space       & CycleGAN           & EMO-DB/SAVEE/IEMOCAP & ACC: 86.32/73.82/61.67 \\ \hline
Yi and Mak \cite{yi2020improving}             & 2020 & Feature Space       & GAN                & EMO-DB/IEMOCAP       & ACC: 84.49/71.45       \\ \hline
Latif et al. \cite{latif2020augmenting}       & 2020 & Feature Space       & GAN                & IEMOCAP/MSP-IMPROV   & UAR: 59.6/46.60        \\ \hline
Sahu et al. \cite{sahu2018enhancing}          & 2018 & Feature Space       & GAN + CGAN & IEMOCAP              & UAR: 60.29             \\ \hline
He et al. \cite{he2020image2audio} &
  2020 &
  \begin{tabular}[c]{@{}l@{}}Sample Space \\ (Cross-modality)\end{tabular} &
  GAN + VAE &
  \begin{tabular}[c]{@{}l@{}}Multi-PIE/CK+/MMI/\\ Oulu-CASIA\end{tabular} &
  ACC: 61.3 \\ \hline
Ma et al. \cite{Ma2023} &
  2023 &
  \begin{tabular}[c]{@{}l@{}}Sample Space \\ (Cross-modality)\end{tabular} &
  GPT-4 &
  IEMOCAP &
  ACC: 68.85 \\ \hline
Malik et al. \cite{Malik2023} &
  2023 &
  \begin{tabular}[c]{@{}l@{}}Sample Space \\ (Cross-modality)\end{tabular} &
  DM &
  \begin{tabular}[c]{@{}l@{}}IEMOCAP/MSP-IMPROV/\\ CREMA-D/RAVDESS\end{tabular} &
  UAR: 61.22 \\ \hline
\end{tabular}%
}
\end{table}

\subsection{Speech Feature Extraction}
\label{5.2}
\quad 
Feature extraction plays a vital role in SER, as it involves deriving meaningful representations from speech signals to capture emotional cues effectively. Traditionally, handcrafted features such as Mel-Frequency Cepstral Coefficients (MFCCs) \cite{davis1980comparison} and prosodic features have been extensively utilized in SER systems. However, these handcrafted features have limitations in capturing subtle emotional variations and can be sensitive to noise and speaker variability \cite{mustaqeem2019cnn}. 
To address these limitations, generative models, have emerged as a promising approach to learn and extract expressive representations directly from the raw speech signals \cite{latif2021survey}. Table \ref{fe ser} provides an overview of the literature on feature extraction and other applications in SER.

\begin{figure}[]
  \centering
  \includegraphics[width=0.8\textwidth]{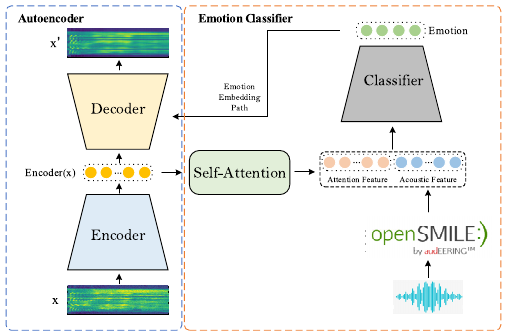}
  \caption{
  The framework designed for extracting speech features from \cite{Zhang2021}: 
  It introduces instance normalization and an emotion embedding path to guide the AE in learning a priori knowledge from the label, enabling it to distinguish the most emotion-related features. 
  The latent representation learned by the AE, enhanced with self-attention, is concatenated with acoustic features obtained using the openSMILE toolkit, and the resulting feature vector is then used for emotion classification.
  }
  \label{speech feature extration}
\end{figure}

For example,
Latif et al. \cite{Latif2017} employ VAEs to learn latent representations of speech emotion and utilize Long Short-Term Memory (LSTM) networks for classifying phonological emotions. 
Zhang and Xue \cite{Zhang2021} propose a hybrid approach that combines the potential representations learned by an AE with acoustic features obtained using the openSMILE toolkit, as shown in Figure \ref{speech feature extration}. The concatenated feature vectors are then used for emotion classification. 
Sahu et al. \cite{sahu2018adversarial} 
leverage an Adversarial AE (AAE) framework to learn low-dimensional code vectors that retain class discriminability from the higher-dimensional feature space for SER.
Ying et al. \cite{ying2021unsupervised} employ Denoising AEs (DAEs) and AAEs in conjunction with unsupervised learning to extract features for model pre-training. 
Almotlak et al. \cite{almotlak2020variational} extend the VAE framework to disentangle speaker-dependent information, such as identity and gender, from speaker-independent features like emotions, providing a more nuanced representation of speech data. 
Fonnegra and Díaz \cite{fonnegra2018speech} use paralinguistic features and a deep convolutional Stacked AE (SAE) network to transform a set of 1582 INTERSPEECH 2010 features \cite{schuller2010interspeech} extracted from speech signals into a higher-level representation for SER.
Sahu et al. \cite{Sahu2022} use two methods to generate emotion-specific feature vectors with GAN variants: training a generator using samples from a mixture prior and providing a one-hot emotion vector to the generator to explicitly generate features for the specified emotion.

\begin{landscape}
\begin{table}[]
\centering
\caption{Literature on Generative Models for Feature Extraction, Semi-supervised learning, Cross-domain, in SER.}
\label{fe ser}
\resizebox{\columnwidth}{!}{%
\begin{tabular}{|l|l|l|l|l|l|}
\hline
Reference                                      & Year & Model     & Application              & Dataset                 & Performance (\%) \\ \hline
Latif et al. \cite{Latif2017}                  & 2017 & VAE       & Feature Extraction       & IEMOCAP                 & ACC: 64.93      \\ \hline
Zhang and Xue \cite{Zhang2021}                 & 2021 & AE        & Feature Extraction       & EmoDB/IEMOCAP           & ACC: 95.6/71.2  \\ \hline
Sahu et al. \cite{sahu2018adversarial}         & 2018 & AAE       & Feature Extraction       & IEMOCAP                 & UAR: 58.38      \\ \hline
Ying et al. \cite{ying2021unsupervised}        & 2021 & DAE + AAE & Feature Extraction       & IEMOCAP                 & ACC: 76.89      \\ \hline
Almotlak et al. \cite{almotlak2020variational} & 2020 & VAE       & Feature Extraction       & OMG                     & ACC: 99.86      \\ \hline
Fonnegra and Díaz \cite{fonnegra2018speech} & 2018 & DCAE     & Feature Extraction       & eNTERFACE’05       & ACC: 91.4        \\ \hline
Sahu et al. \cite{Sahu2022}                    & 2022 & GAN + AE  & Feature Extraction       & IEMOCAP/MSP-IMPROV      & ACC: 45.56      \\ \hline
Zhao et al. \cite{Zhao2020}                    & 2020 & GAN       & Semi-Supervised Learning & IEMOCAP                 & UAR: 58.7       \\ \hline
Chang and Scherer \cite{chang2017learning}  & 2017 & DCGAN    & Semi-Supervised Learning & IEMOCAP            & ACC: 49.80       \\ \hline
Deng et al. \cite{deng2017semisupervised}      & 2017 & AE        & Semi-Supervised Learning & FAU Aibo                & UAR: 63.6       \\ \hline
Xiao et al. \cite{Xiao2023}                    & 2023 & VAE       & Semi-Supervised Learning & FAU Aibo                & UAR: 42.9       \\ \hline
Neumann and Vu \cite{neumann2019improving}  & 2019 & AE       & Semi-Supervised Learning & IEMOCAP/MSP-IMPROV & UAR: 59.54/45.76 \\ \hline
Latif et al. \cite{latif2020multi}          & 2020 & AAE      & Semi-Supervised Learning & IEMOCAP/MSP-IMPROV & ACC: 66.7/60.2   \\ \hline
Zhou et al. \cite{zhou2018inferring}           & 2018 & VAE       & Semi-Supervised Learning & IEMOCAP                 & ACC: 76.7       \\ \hline
Nasersharif et al. \cite{Nasersharif2023}      & 2023 & AE        & Cross-domain SER         & EMODB/EMOVO             & ACC: 57.71      \\ \hline
Xiao et al. \cite{Xiao2020}                    & 2020 & VAE       & Cross-domain SER         & IEMOCAP/MSP-IMPROV      & UAR: 44.1/56.2  \\ \hline
Das et al. \cite{Das2022}                      & 2022 & VAE       & Cross-domain SER         & IEMOCAP                 & ACC: 52.09      \\ \hline
Latif et al. \cite{latif2022self}              & 2022 & ADDi      & Cross-domain SER         & EmoDB/IEMOCAP           & UAR: 47.1/49.8  \\ \hline
Su et al. \cite{Su2023}                        & 2023 & CycleGAN  & Cross-domain SER         & IEMOCAP/VAM/MSP-Podcast & ACC: 61.64      \\ \hline
Su and Lee \cite{su2021conditional}         & 2021 & CycleGAN & Cross-domain SER         & IEMOCAP/MSP-IMPROV & ACC: 51.13/65.7  \\ \hline
\end{tabular}%
}
\end{table}
\end{landscape}

\subsection{Semi-Supervised Learning}
\label{5.3}
\quad 
Unlabeled data refers to samples that lack corresponding labels, which may be due to the failure to obtain appropriate labels during the data collection process or the inability to cover the entire dataset due to the high cost of manual labeling. 
In the field of emotion recognition, the presence of unlabeled data can hinder the performance of models in emotion classification tasks \cite{parthasarathy2020semi}. 
To address this issue, semi-supervised learning emerges as a promising approach that leverages both labeled and unlabeled data to improve the generalization and performance of emotion recognition models \cite{Zhao2020,deng2017semisupervised}.
Common semi-supervised learning methods include strategies based on label propagation, self-training, and generative models \cite{van2020survey}. 
Among these approaches, generative models have proven to be particularly effective.

\begin{figure}[t]
  \centering
  \includegraphics[width=0.9\textwidth]{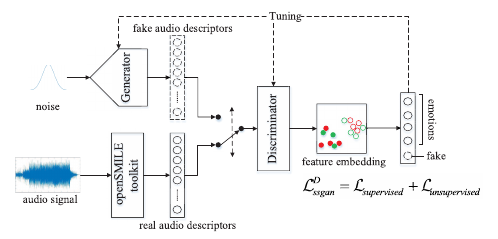}
  \caption{A semi-supervised learning framework in SER from \cite{Zhao2020}: A generator creates synthetic audio descriptors from noise. These descriptors, along with real ones from the openSMILE toolkit, are fed to a discriminator. The discriminator is trained with supervised and unsupervised loss functions to better distinguish real from fake audio cues.}
  \label{semi speech}
\end{figure}

To be specific, 
Zhao et al. \cite{Zhao2020} propose a semi-supervised GAN for SER, which is designed to capture underlying knowledge from both labeled and unlabeled data, as shown in Figure \ref{semi speech}. 
In their approach, a generator creates synthetic audio descriptors from noise, while a discriminator is trained to distinguish between real and fake audio cues using both supervised and unsupervised loss functions. The discriminator not only classifies input samples as real or fake but also learns to identify the emotional class of real samples.
Chang and Scherer \cite{chang2017learning} present a multitask deep convolutional GAN (DCGAN) approach for emotional valence classification in speech, which leverages a multi-task annotated corpus and a large unlabeled conference corpus. 
In addition to GANs, AEs have also been widely used by researchers for semi-supervised learning in SER. 
For example, 
Deng et al. \cite{deng2017semisupervised} propose a semi-supervised AE model for SER that combines both generative and discriminative perspectives, enabling the model to distill knowledge from unlabeled data and incorporate it into the supervised learning process, thus enhancing the overall performance of the SER system.
Xiao et al. \cite{Xiao2023} propose a novel semi-supervised adversarial VAE that learns the distribution of shared hidden features of labeled and unlabeled data. 
Neumann and Vu \cite{neumann2019improving} employ a recurrent sequence-to-sequence AE trained on unlabeled data to generate representations for the labeled data. These generated representations serve as auxiliary information to the classifier during training, helping to improve the emotion recognition performance.
Latif et al. \cite{latif2020multi} design an unsupervised AAE and combine it with a supervised classification network to achieve semi-supervised learning. 
Zhou et al. \cite{zhou2018inferring} extend the multi-path deep neural network to a generative model based on semi-supervised VAE, which enables simultaneous training on both labeled and unlabeled data.

\begin{figure}[t]
  \centering
  \includegraphics[width=1\textwidth]{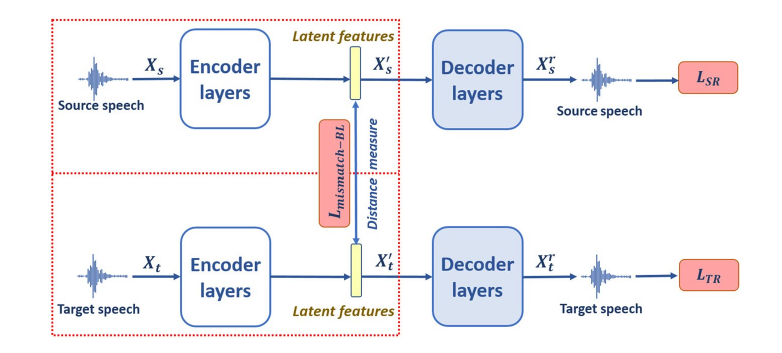}
  \caption{
  A framework \cite{Nasersharif2023} which leverages AEs to address the cross-domain problem in SER: The key idea is to utilize the latent representations learned by the AE to align and compare the features of the source and target speech domains.
  }
  \label{speech cross domain}
\end{figure}

\subsection{Cross-domain SER}
\label{cdser}
Here, we discuss cross-domain SER, which refers to the ability of a model trained on one speech emotion domain (source domain) to be extended and applied to another different speech emotion domain (target domain). 
This cross-domain generalization is a major challenge faced by SER because different speech emotion domains may have significant differences, such as language, culture, speaking style, recording conditions, and so on \cite{redko2020survey,farahani2021brief}. 
These differences form distribution inconsistencies between different domains, so that the performance of a model trained on one domain often decreases significantly on another domain. 
Generative models provide a new perspective for solving this problem. 
By introducing domain adaptation techniques, generative models can learn to map samples from the source dataset and target dataset to a shared feature space, thereby eliminating the distribution differences between them.

For example, 
Nasersharif et al. \cite{Nasersharif2023} employ separate AEs for each source and target domain dataset and introduce a domain adaptation loss in addition to the conventional loss to achieve domain-invariant feature extraction, as shown in Figure \ref{speech cross domain}. 
Xiao et al. \cite{Xiao2020} design a framework by incorporating a VAE to extract generalized and domain-invariant representations from latent feature distributions. 
Das et al. \cite{Das2022} propose a VAE-based method with KL annealing and semi-supervised learning to achieve more consistent latent embedding distributions across datasets. 
Latif et al. \cite{latif2022self} introduce an Adversarial Dual Discriminator (ADDi) network, which generates domain-invariant representations through a three-player adversarial game between a generator and two discriminators.
Su et al. \cite{Su2023} introduce corpus-aware emotional CycleGAN, which incorporates a corpus-aware attention mechanism to aggregate each source dataset and generate synthetic target samples.
Su and Lee \cite{su2021conditional} propose a conditional cycle GAN to generate samples similar to the source domain and assign the original labels of the source samples.

\subsection{Adversarial Sample Generation and Defense}
\label{5.4}
\quad 
Adversarial examples refer to samples that are created by adding carefully designed perturbations to the original samples, causing the model to misclassify them \cite{goodfellow2014explaining}. In the SER task, adversarial examples can be speech with specific noise or perturbations added, which can deceive the SER model and lead to incorrect emotion predictions \cite{Gong2017}. The existence of adversarial examples highlights the vulnerability of SER models, as these models may overly rely on certain specific acoustic features while ignoring the overall features of emotional expression. Adversarial example defense refers to improving the ability of SER models to resist adversarial attacks, enabling them to maintain high emotion recognition accuracy even when faced with adversarial examples.

Generative models can play an important role in the generation and defense of adversarial examples in SER. 
On the one hand, generative models can be used to generate realistic adversarial examples. 
For example, Chang et al. \cite{chang2024staa} introduce STAANet, a novel method based on WaveNet \cite{van2016wavenet} for generating sparse and transferable adversarial examples to spoof SER systems. 
Wang et al. \cite{Wang2020} propose a GAN-based attack approach that can efficiently generate adversarial examples with pre-specified target labels.
On the other hand, generative models can also be used to construct adversarial example defense systems.
For example, Latif et al. \cite{Latif2018} introduce a two-step defense strategy for SER systems. In their approach, perturbed utterances are first cleaned using GANs, and then a classifier is applied to the cleaned data. By removing the adversarial perturbations from the input samples, the classifier can make more accurate predictions and maintain robust performance.
Chang et al. \cite{Chang2022} propose a framework based on federated adversarial learning to protect data privacy and defend against adversarial attacks in SER systems. Their approach leverages the distributed nature of federated learning to train models on decentralized data while incorporating adversarial training techniques to enhance robustness against adversarial examples.

\vspace{1em}
\quad In summary, 
generative models have made significant contributions to SER by enabling data augmentation, feature extraction, semi-supervised learning, and adversarial sample generation and defense. 
These applications have greatly enhanced the performance of SER systems. 
As research in this area continues to advance, generative models are expected to play an increasingly crucial role in pushing the boundaries of SER.

\section{Facial Emotion Recognition (FER)}
\label{fer}
\quad 
FER is a research field that utilizes computer vision techniques to recognize and understand the emotional states of individuals by analyzing facial expressions in facial images or video frames 
\cite{canal2022survey,li2020deep}. In recent years, generative models have made impressive progress in the field of FER, which is similar to SER, but presents some unique research focuses and development directions.
This section will provide a detailed discussion on the applications of generative models in FER. 
Firstly, Section \ref{6.1} will discuss facial data augmentation. 
Secondly, Section \ref{6.2} will focus on facial feature extraction. 
Next, Section \ref{6.3} will introduce semi-supervised learning for FER. 
Furthermore, Section \ref{6.4} will focus on FER in typical noisy environments. 
Finally, Section \ref{6.5} will discuss dynamic FER.

\subsection{Facial Data Augmentation}
\label{6.1}
\quad 
Similar to speech data, acquiring high-quality facial image data is an expensive and time-consuming task, which hinders the development of FER \cite{mollahosseini2017affectnet,dalvi2021survey}. 
With generative models, it becomes possible to generate synthetic facial images that align with the desired emotion labels, thus providing a data augmentation method to overcome the scarcity of labeled data.
Table \ref{da fer} shows the literature on generative models for data augmentation in FER.

\begin{figure}[t]
  \centering
  \includegraphics[width=0.85\textwidth]{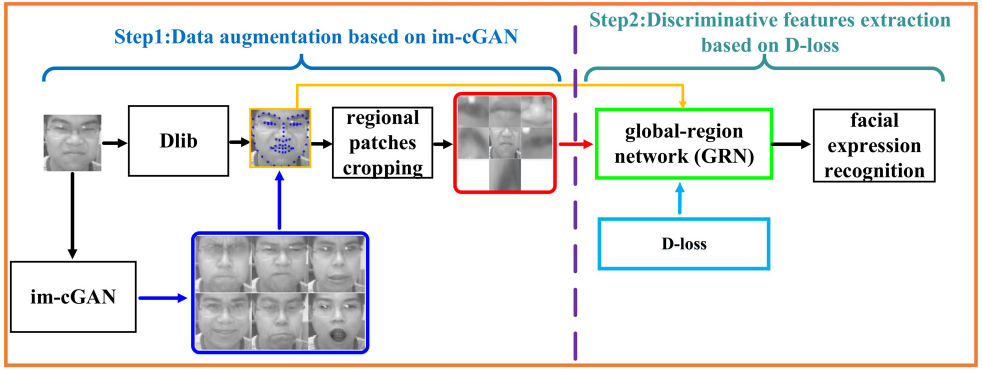}
  \caption{
   A facial data generation framework based on GANs from \cite{Sun2023}: The original image undergoes facial landmarks detection using Dlib, after which the im-cGAN generates augmented images and regional patches. These patches are then cropped into specific regions. In the second step, discriminative features are extracted. The cropped regional patches are fed into a global-region network, which extracts discriminative features based on the D-loss function. 
  }
  \label{face data augmentation}
\end{figure}

Specifically, a series of GAN-based methods are proposed.
Porcu et al. \cite{porcu2020evaluation} reveal that GAN-based data augmentation substantially outperforms geometric transformations in improving emotion recognition accuracy of a VGG16 CNN-based FER system \cite{Simonyan15}, highlighting the significance of large training datasets and the potential of combining GAN with less complex techniques to mitigate computational complexity.
Zhu et al. \cite{zhu2018emotion} propose a data augmentation method using a framework combining a CNN classifier and a CycleGAN generator with least-squared adversarial loss to complement and complete the data manifold, find better margins between neighboring classes, and avoid the gradient vanishing problem.
As shown in Figure \ref{face data augmentation}, Sun et al. \cite{Sun2023} propose an improved CGAN (im-cGAN) model trained on facial images with Action Units (AUs) to generate more labeled samples for data augmentation. 
Similarly, Wang et al. \cite{Wang2019} introduce an end-to-end synthetic Compositional GAN (Comp-GAN) that employs six task-driven loss functions to generate more realistic and natural facial images. 
Kusunose et al. \cite{Kusunose2022} propose a data augmentation method using StyleGAN2 \cite{karras2020analyzing} to generate artificial facial expression images for seven emotions, which are used as additional training data to train a VGG16-based emotion recognition model through transfer learning.
Yang et al. \cite{YANG2022} introduce an end-to-end deep learning framework for generating high-quality facial images. The framework is built upon DCGAN and incorporates an Ensemble Networks (Ens-Net) model for network integration.
Han et al. \cite{Han2023} improve upon the image conversion model StarGAN V2 \cite{Szegedy2015} by integrating the Squeeze-and-Excitation Network (SENet) \cite{hu2018squeeze} into the generator. They also introduce hinge losses \cite{liu2020saanet} to enhance the authenticity of the generated images. These modifications aim to produce more realistic and diverse facial expressions across multiple domains.
Wang et al. \cite{wang2020laun} propose an improvement to the StarGAN model for data augmentation in FER by introducing context loss and an attention mechanism. These modifications enable the model to generate more realistic and expressive facial images with fine details in crucial regions, such as eyes and mouth. 

In addition to GAN-based methods, there are also some approaches that combine AEs for data augmentation in FER. For example, 
Li et al. \cite{li2023semantic} introduce a novel approach that integrates GAN and VAE in the latent space for data augmentation in FER. The proposed method aims to generate diverse facial images that carry rich semantic information, enhancing the quality and variety of the augmented dataset.
Pons et al. \cite{pons2020cyclegan} propose using CycleGAN to generate thermal facial images from visual spectrum images, thereby augmenting the training dataset for thermal emotion recognition. This can be seen as a cross-modality data augmentation approach.

\begin{table}[]
\centering
\caption{Literature on Generative Models for Data Augmentation in FER.}
\label{da fer}
\resizebox{\columnwidth}{!}{%
\begin{tabular}{|l|l|l|l|l|}
\hline
Reference                               & Year & Model     & Dataset   & Performance (\%) \\ \hline
Porcu et al. \cite{porcu2020evaluation} & 2022 & GAN       & CK+       & ACC: 83.3        \\ \hline
Zhu et al. \cite{zhu2018emotion}        & 2018 & CycleGAN  & FER2013   & ACC: 94.65       \\ \hline
Sun et al. \cite{Sun2023}       & 2023 & CGAN    & JAFFE/CK+/Oulu-CASIA/KDEF & ACC: 98.37/98.10/93.34/98.30 \\ \hline
Wang et al. \cite{Wang2019}             & 2019 & GAN       & F2ED      & ACC: 44.19       \\ \hline
Kusunose et al. \cite{Kusunose2022}     & 2022 & StyleGAN2 & CFEE      & ACC: 82.04       \\ \hline
Yang et al. \cite{YANG2022}     & 2022 & GAN     & FER2013/CK+/JAFFE         & ACC: 82.1/84.8/91.5          \\ \hline
Han et al. \cite{Han2023}               & 2023 & starGAN   & CK+/MMI   & ACC: 99.20/98.14 \\ \hline
Wang et al. \cite{wang2020laun} & 2020 & starGAN & KDEF/MMI                  & ACC: 95.97/98.30             \\ \hline
Li et al. \cite{li2023semantic}         & 2023 & GAN + VAE & RAF-DB    & ACC: 74.43       \\ \hline
Pons et al. \cite{pons2020cyclegan}     & 2020 & CycleGAN  & USTC-NVIE & ACC: 51          \\ \hline
\end{tabular}%
}
\end{table}

\subsection{Facial Feature Extraction}
\label{6.2}
\quad 
Traditional FER methods often rely on handcrafted features or predefined facial landmarks to describe facial expressions \cite{canal2022survey,li2020deep}.
These features typically include the locations of facial key points, geometric relationships, and local texture descriptors \cite{ekman1978facial}. 
However, due to the complexity and diversity of facial expressions, 
these handcrafted features may not adequately capture the subtle differences between different expressions \cite{minaee2021deep}. 
Moreover, the variations in facial features among different individuals also pose challenges to feature design \cite{ko2018brief,li2018occlusion}. 
In recent years, with the development of deep learning, generative models provide new ideas for solving the feature extraction problem in FER by automatically learning compact and informative representations of facial expressions. 
Table \ref{fe fer} shows a series of research works,
where
CCC (Concordance Correlation Coefficient) 
assesses the agreement between predicted and actual values, taking into account both the correlation and the deviation from perfect agreement.

\begin{figure}[t]
  \centering
  \includegraphics[width=0.8\textwidth]{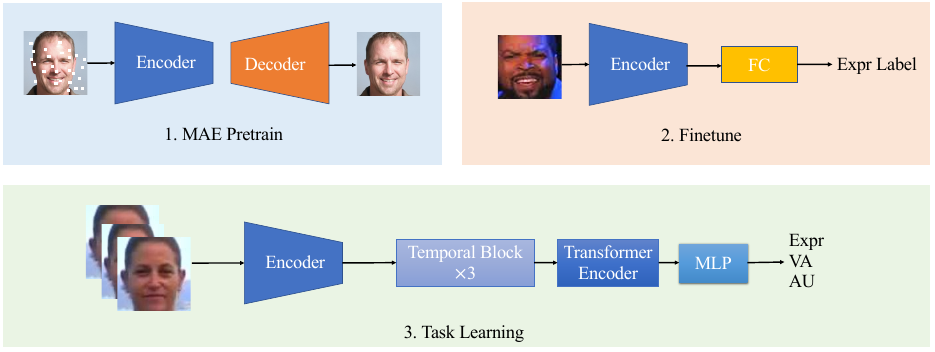}
  \caption{
  A Masked Autoencoder (MAE)-based facial feature extraction framework from \cite{zhou2024enhancing}: 
  It consists of a pre-trained ViT model using MAE on facial expression datasets, followed by fine-tuning the encoder with a fully connected layer and expression labels, and finally connecting the fine-tuned encoder with temporal blocks, a Transformer encoder, and a MLP for sequence task learning.
  }
  \label{facial_feature_extraction}
\end{figure}

For example, 
Yang et al. \cite{Yang2020} propose the Exchange-GAN model, which achieves the separation of expression-related and expression-irrelevant features through partial feature exchange and various constraints, such as adversarial loss, classification loss, content loss, and central loss. As illustrated in Figure \ref{facial_feature_extraction}, this approach effectively disentangles the facial expression information from identity-related features.
Khemakhem and Ltifi \cite{Khemakhem2023} introduce a neural style transfer GAN (NST-GAN) to remove identity information from facial images. By converting each image into a synthetic ``average'' identity, NST-GAN focuses on extracting expression-specific features while minimizing the influence of individual variations.
Analogously, Yang et al. \cite{yang2018facial} propose a de-expression residue learning method by training a CGAN to generate neutral face images from input expressions, and then learn the expressive information residue from the intermediate layers of the generative model.
Zhang and Tang \cite{zhang2021efficacious} use a GAN to generate a neutral face from an emotional face, extract features from both faces using separate convolutional layers, and then subtract the neutral features from the emotional features to obtain pure ``emotion features" for FER.
Xie et al. \cite{Xie2021} propose the two-branch divergent GAN, which consists of two generators and discriminators. One branch is dedicated to facial expression feature extraction and classification, while the other focuses on facial feature extraction and identity discrimination. This architecture allows for a more targeted approach to capturing expression-related information.
Ali and Hughes \cite{ali2019facial} propose a novel disentangled expression learning GAN that decouples expression representation from identity components. By explicitly providing identity codes to the decoder, this approach enables the model to learn expression features independently of subject-specific characteristics.
Similarly, 
Tiwary et al. \cite{Tiwary2023} 
design an expression GAN to separate shape, skin color, and other identity-related information from specific muscle movements that are crucial for emotion recognition.
Sima et al. \cite{sima2021automatic} tackle the challenges of category overlap and facial expression variation due to individual differences using CGAN. By generating neutral expressions while retaining identity-related information, their method aims to normalize the facial images and facilitate more accurate expression recognition.
Similar GAN-based approaches for addressing facial feature extraction include \cite{yang2018identity,wang2021improved,abiram2021identity,dharanya2021facial}.

In addition to the aforementioned methods based on GANs, several AE-based approaches are proposed to address the feature extraction problem in FER. 
For example,
Kim et al. \cite{kim2017deep} design an AE network with CNNs to learn a contrastive representation of facial expressions by comparing a query image with a generative reference image, estimated from the given image itself. 
Wu et al. \cite{Wu2020} 
propose Cross-VAE, an extension of conditional VAE that disentangles expression from identity by assuming orthogonal latent representations and employing a symmetric training procedure, ensuring disentangled and expressive encodings.
Chatterjee et al. \cite{chatterjee2022improving} propose a residual VAE model to significantly reduce class overlapping in the feature space by transforming facial images into latent vector. 
They \cite{chatterjee2024majority} also apply various resampling techniques after feature extraction to solve the class imbalance problem.
Zhou et al. \cite{zhou2024enhancing} use Masked Autoencoders (MAEs) for model pre-training and improve the ability to extract facial features through masked-reconstruction method, as shown in Figure \ref{facial_feature_extraction}.

\begin{table}[]
\centering
\caption{Literature on Generative Models for Feature Extraction in FER.}
\label{fe fer}
\resizebox{\columnwidth}{!}{%
\begin{tabular}{|l|l|l|l|l|}
\hline
Reference                                        & Year & Model    & Dataset                    & Performance (\%)            \\ \hline
Yang et al. \cite{Yang2020}                      & 2020 & GAN      & Multi-PIE/FACES/Oulu-CASIA & ACC: 91.08/95.24/86.33      \\ \hline
Khemakhem and Ltifi \cite{Khemakhem2023}         & 2023 & GAN      & CK+/JAFFE/FER 2013         & ACC: 98.14/95.12/85.93      \\ \hline
Yang et al. \cite{yang2018facial}                & 2018 & CGAN     & CK+/MMI/BU-3DFE/BP4D+      & ACC: 97.30/88.0/73.23/84.17 \\ \hline
Zhang and Tang \cite{zhang2021efficacious}       & 2021 & GAN      & CK+                        & ACC: 98.04                  \\ \hline
Xie et al. \cite{Xie2021}                        & 2021 & GAN    & CK+/TFEID                  & ACC: 97.53/97.20            \\ \hline
Ali and Hughes \cite{ali2019facial}              & 2019 & GAN      & CK+/Oulu-CASIA /MMI        & ACC: 97.28/89.17/72.97      \\ \hline
Tiwary et al. \cite{Tiwary2023}                  & 2023 & GAN    & CK+/Oulu-CASIA/FER 2013    & ACC: 94.67/92.57/92.85      \\ \hline
Sima et al. \cite{sima2021automatic}             & 2021 & CGAN     & CK/MMI                     & ACC: 93.52/93.60            \\ \hline
Yang et al. \cite{yang2018identity} & 2018 & CGAN & CK+/Oulu-CASIA/BU-3DFE/BU-4DFE & ACC: 96.57/88.92/76.83/89.55 \\ \hline
Wang \cite{wang2021improved}                     & 2021 & GAN      & JAFFE/CK+/FER 2013         & ACC: 96.6/95.6/72.8         \\ \hline
Abiram et al. \cite{abiram2021identity}          & 2021 & StyleGAN & CK+                        & ACC: 96.97                  \\ \hline
Dharanya et al. \cite{dharanya2021facial}        & 2021 & ACGAN    & ADFES-BIV/CK+/KDEF         & ACC: 93.6/97.39/96.33       \\ \hline
Kim et al. \cite{kim2017deep}                    & 2017 & AE       & CK+/MMI/Oulu-CASIA         & ACC: 97.93/81.53/86.11      \\ \hline
Wu et al. \cite{Wu2020}                          & 2020 & VAE      & CK+/Oulu-CASIA/RAF-DB      & ACC: 94.96/86.87/84.81      \\ \hline
Chatterjee et al. \cite{chatterjee2022improving} & 2022 & VAE      & Affectnet                  & ACC: 95                     \\ \hline
Zhou et al. \cite{zhou2024enhancing}             & 2024 & MAE      & Affwild                    & CCC: 60.57                  \\ \hline
\end{tabular}%
}
\end{table}

\subsection{Cross-domain FER}
\label{6.5}
\quad

Similar to cross-dataset SER, generative models can also provide effective solutions for cross-dataset FER through domain adaptation.
For instance,
Wang et al. \cite{wang2018unsupervised} propose an unsupervised domain adaptation method based on GANs. Their approach generates new samples to fine-tune the pre-trained FER model, enabling it to better adapt to the target domain and achieve higher accuracy across different datasets.
Fan et al. \cite{fan2018unsupervised} employ CGANs to learn domain-invariant feature representations of facial expressions, allowing for effective knowledge transfer from labeled source domains to unlabeled target domains.
Zhang et al. \cite{zhang2018facial} propose a domain adaptation model for FER, which utilizes unlabeled web facial images as an auxiliary domain to generate labeled facial images in the target domain using GANs with an attention transfer module, preserving structural consistency through pixel cycle-consistency and discriminative loss, and leveraging attention maps from the source domain classifier to enhance the target domain classifier's performance.

\subsection{FER in Noisy Environments}
\label{6.3}
\quad 
In real-world applications, facial images are often affected by various noisy environments, such as occlusions, complex backgrounds, non-frontal views, and other factors that reduce the performance of FER \cite{li2020deep}. In recent years, generative models demonstrate powerful capabilities in effectively addressing these noisy environments.
Table \ref{tab:my-table} presents the literature on generative models for FER in noisy environments.

For example, to cope with occlusion and complex backgrounds,
Du et al. \cite{du2021learning} introduce an expression associative network that learns the associations between inter- and intra-class expressions, along with a GAN-based auxiliary module designed to suppress changes caused by occlusion, illumination, and pose variations. 
Peng et al. \cite{peng2020emotion} propose a method based on ACGAN \cite{odena2017conditional} to restore occluded facial images, thereby enhancing the data manifold and improving FER performance.
Similarly, Lu et al. \cite{lu2019wgan} employ a Wasserstein GAN-based method to generate unoccluded facial images, enabling effective FER even in the presence of partial occlusion.
To mitigate the influence of background clutter on FER, Tang et al. \cite{tang2019expression} introduce an expression CGAN that incorporates a novel mask loss. This loss function helps to reduce the impact of background information, allowing the model to focus more on the facial regions relevant to expression recognition.
Building upon the idea of masking, Li et al. \cite{Li2021} propose a mask generation network (MGN) inspired by GANs. The MGN effectively filters out background noise and interference from facial images by generating masks that cover only the key areas of each face required for accurate expression classification. This targeted masking approach preserves the most informative regions while suppressing irrelevant information.

Apart from occlusion and complex backgrounds, the presence of non-frontal facial images in wild datasets poses another significant challenge for FER due to variations in facial posture. 
To address this issue, researchers have proposed different solutions based on generative models, aiming to generate high-quality frontal-face images from non-frontal views.
For example,
Han and He \cite{Han2021} introduce a GAN-based three-stage training algorithm for multi-view FER. 
They utilize a pre-trained classification model to construct a new GAN that effectively synthesizes frontal faces while preserving facial expression features. 
This approach ensures that the generated frontal images retain the emotional information necessary for accurate FER.
Instead of explicitly generating frontal faces, Zhang et al. \cite{zhang2020geometry} propose a GAN-based method that leverages the geometric information inherent in facial shapes to achieve pose-invariant FER. This innovative perspective offers an alternative solution to the challenge of non-frontal images in wild datasets.
Similar GAN-based methods for non-frontal FER also include \cite{lai2018emotion,li2019multi,Dong2023}.

\begin{figure}[t]
  \centering
  \includegraphics[width=0.9\textwidth]{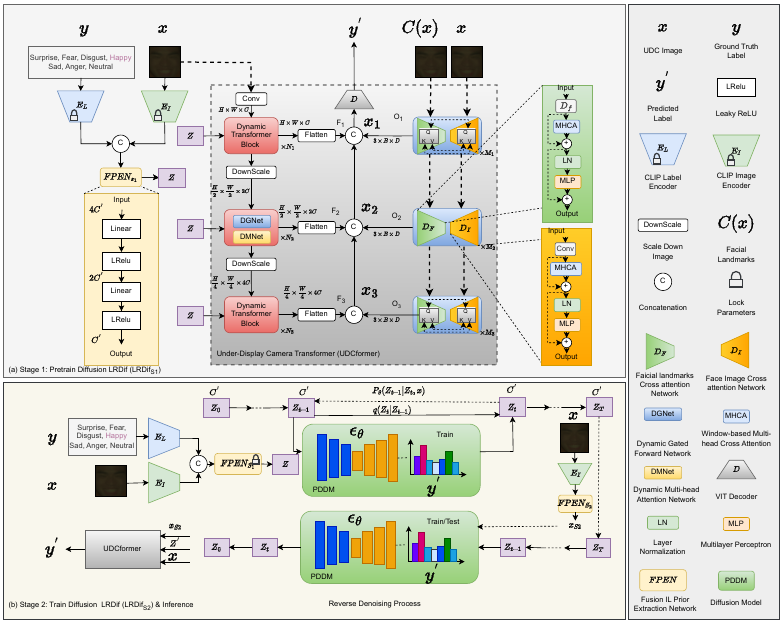}
  \caption{
  A DM-based framework for FER in noisy environments proposed by \cite{wang2024lrdif}: 
  It overcomes image degradation  by under-display cameras (UDC) through a two-stage training strategy involving a preliminary extraction network (FPEN) and a transformer network (UDCformer), enabling effective recovery of emotion labels from degraded UDC images.
  }
  \label{Intensity enhancement}
\end{figure}

Moreover, generative models can be employed to combat other noise factors that degrade the performance of FER.
For example, Yang et al. \cite{Yang2021} propose a GAN-based method for recognizing expressions in low-intensity facial images. The concatenated training critic combines both the general adversarial loss, which regulates the intensity of facial expressions, and the FER penalty, which directs the model to generate visually enhanced facial images.
Nan et al. \cite{Nan2022} propose a novel GAN-based feature-level super-resolution method for robust FER that reduces the risk of privacy leakage. This approach transforms low-resolution image features into more discriminative ones using a pre-trained FER model as a feature extractor. 
To address the issue of image degradation caused by under-display cameras (UDC), Wang et al. \cite{wang2024lrdif} employ a method based on DM to overcome inherent noise and distortion obstacles, thereby improving the accuracy of expression recognition, as shown in Figure \ref{Intensity enhancement}.

\begin{table}[]
\centering
\caption{Literature on Generative Models for FER in Noisy Environments.}
\label{tab:my-table}
\resizebox{\columnwidth}{!}{%
\begin{tabular}{|l|l|l|l|l|l|}
\hline
Type                               & Reference                             & Year & Model           & Dataset            & Performance (\%) \\ \hline
\multirow{5}{*}{\begin{tabular}[c]{@{}l@{}}Occlusion and \\ Complex Backgrounds\end{tabular}} &
  Du et al. \cite{du2021learning} &
  2021 &
  GAN &
  RAF-DB/FER2013Plus &
  ACC: 87.03/90.07 \\ \cline{2-6} 
                                   & Peng et al. \cite{peng2020emotion}    & 2020 & ACGAN           & FER2013            & ACC: 73          \\ \cline{2-6} 
                                   & Lu et al. \cite{lu2019wgan}           & 2019 & Wasserstein GAN & RAF-DB/AffectNet   & ACC: 83.49/59.73 \\ \cline{2-6} 
                                   & Tang et al. \cite{tang2019expression} & 2019 & CGAN            & JAFFE              & ACC: 80.32       \\ \cline{2-6} 
                                   & Li et al. \cite{Li2021}               & 2021 & GAN             & RAF-DB/FER2013Plus & ACC: 87.91/88.88 \\ \hline
\multirow{5}{*}{Non-frontal Views} & Han and He \cite{Han2021}             & 2021 & GAN             & KDEF/Multi-PIE     & ACC: 90.42/92.63 \\ \cline{2-6} 
                                   & Zhang et al. \cite{zhang2020geometry} & 2020 & GAN             & Multi-PIE/BU-3DFE  & ACC: 92.09/81.95 \\ \cline{2-6} 
                                   & Lai and Lai \cite{lai2018emotion}     & 2018 & GAN             & Multi-PIE          & ACC: 86.76       \\ \cline{2-6} 
                                   & Li et al. \cite{li2019multi}          & 2019 & GAN             & Multi-PIE          & ACC: 86.9        \\ \cline{2-6} 
                                   & Dong et al. \cite{Dong2023}           & 2023 & GAN             & KDEF               & ACC: 92.7        \\ \hline
\multirow{3}{*}{Other Factors}     & Yang et al. \cite{Yang2021}           & 2021 & GAN             & BU-3DFE/BU-4DFE    & ACC: 88.89/80.03 \\ \cline{2-6} 
                                   & Nan et al. \cite{Nan2022}             & 2022 & GAN             & RAF-DB/SFEW2.0     & ACC: 84.09/55.14 \\ \cline{2-6} 
 &
  Wang et al. \cite{wang2024lrdif} &
  2024 &
  DM &
  \begin{tabular}[c]{@{}l@{}}UDC-RAF-DB/\\ UDC-FERPlus\end{tabular} &
  ACC: 88.55/84.89 \\ \hline
\end{tabular}%
}
\end{table}

\subsection{Dynamic FER}
\label{6.4}
\quad 
The previously discussed FER mainly focuses on static images, i.e., single-frame facial expression recognition. However, in real-world scenarios, emotions often change dynamically, requiring consideration of the temporal evolution of facial expressions. 
Dynamic FER aims to analyze the changes in facial expressions across video sequences, capturing the temporal dynamic features of emotions \cite{saleem2021real}. Compared to static FER, dynamic FER can provide richer and more accurate emotional information, but it also presents greater challenges due to the complexity of temporal dynamics and the need for robust algorithms that can handle variations in facial movements, head poses, and lighting conditions across frames \cite{pu2015facial,krumhuber2017review,deng2020mimamo}. 
Table \ref{dynamic fer} shows the literature on generative models for dynamic FER, where UF1 stands for unweighted F1 score that combines precision and recall, providing a balanced measure of a model's performance.

\begin{figure}[t]
  \centering
  \includegraphics[width=1\textwidth]{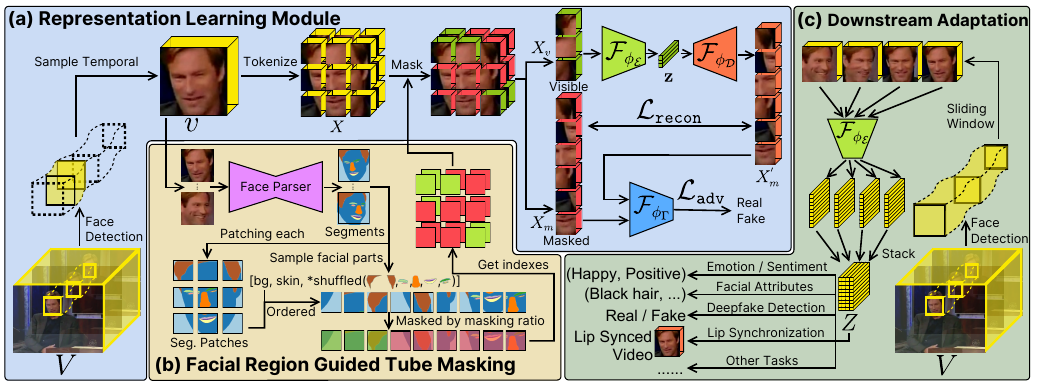}
  \caption{
  A MAE framework for dynamic FER from \cite{cai2023marlin}: 
  It begins by detecting faces in each video frame, followed by a temporal sampling and tokenization process that divides the detected faces into segments. These segments are then fed into a face parser, which identifies and segments various facial features, such as eyes, nose, and mouth. The segmented features are patched together to create a complete representation of the face.
  To enhance the learning process, the framework employs a selective masking strategy. It randomly masks some of the patches, creating a mix of visible and masked segments. 
  }
  \label{dfer}
\end{figure}

For example, 
Cai et al. \cite{cai2023marlin} introduce a Masked Autoencoder (MAE) for learning robust and generic facial embeddings, as depicted in Figure \ref{dfer}. The MAE reconstructs spatio-temporal details of the face from densely masked facial regions, capturing both local and global aspects to encode transferable features. 
Priyanka A. Gavade et al. \cite{Gavade2023} propose the Taylor-Chicken Swarm Optimization-based Deep GAN (Taylor-CSO-based Deep GAN), which provides a new method for recognizing video expressions. The Deep GAN, trained by the Taylor CSO approach, is adopted to efficiently complete the recognition process on the basis of the feature matrix.
Guo et al. \cite{Guo2023} propose a novel approach that leverages GANs to generate inter-class optical flow images for FER. They calculate the difference between static fully expressive samples and neutral expression samples to obtain the optical flow information. By training the GAN on these optical flow images, the model learns to capture the subtle motion patterns associated with different facial expressions. 
Similarly, Liong et al. \cite{liong2020evaluation} propose an improved FER system that utilizes GANs to generate inter-class optical flow images by computing variations of optical flow. They utilize a self-attention mechanism within the GAN architecture, which enables the generator to integrate image information from all feature locations when generating image details.
Other similar works based on GANs for dynamic FER include \cite{chen2021dffcn,mazen2021real}.

\begin{table}[]
\centering
\caption{Literature on Generative Models for Dynamic FER.}
\label{dynamic fer}
\resizebox{\columnwidth}{!}{%
\begin{tabular}{|l|l|l|l|l|}
\hline
Reference                               & Year & Model    & Dataset            & Performance (\%)       \\ \hline
Cai et al. \cite{cai2023marlin}         & 2023 & MAE      & CMU-MOSEI          & ACC: 80.60             \\ \hline
Gavade et al. \cite{Gavade2023}         & 2023 & DGAN     & CK+/RAVDESS/SAVEE  & ACC: 82.14/80/79       \\ \hline
Guo et al. \cite{Guo2023}               & 2023 & GAN      & SAMM/AFEW          & ACC: 56.98/60.20/44.72 \\ \hline
Liong et al. \cite{liong2020evaluation} & 2020 & GAN      & CASME II/SMIC/SAMM & ACC: 70.29             \\ \hline
Li et al. \cite{chen2021dffcn}          & 2021 & GAN      & CASME II/SAMM/SMIC & UF1: 84.52, UAR: 84.65   \\ \hline
Mazen et al. \cite{mazen2021real}       & 2021 & CycleGAN & FER2013            & ACC: 91.76             \\ \hline
\end{tabular}%
}
\end{table}

\vspace{1em}
\quad
In summary, generative models show tremendous advantages in the field of FER. First, generative models can effectively augment training data by generating realistic and diverse new facial images. Second, generative models demonstrate unique advantages in facial feature extraction. Moreover, generative models provide effective solutions for cross-domain FER. When dealing with noisy environments, generative models also exhibit promising performance. Finally, the effectiveness of generative models in dynamic FER is fully validated.
In addition, it is worth noting that apart from these aspects,
generative models can also play a role in semi-supervised scenarios for FER, although there are relatively fewer related works. 
For example, Chen et al. \cite{chen2019emotion} propose an improved classification method based on semi-supervised GANs. They replace the output layer of the traditional unsupervised GAN with a Softmax layer, enabling the model to simultaneously learn from labeled and unlabeled data. 

\section{Textual Emotion Recognition (TER)}
\label{ter}
\quad
TER is a highly focused and challenging research direction in the field of natural language processing \cite{alswaidan2020survey}. 
It is dedicated to automatically identifying and extracting the emotional information contained in textual data, providing insights into the emotional states and subjective attitudes conveyed by humans in their written expressions \cite{chatterjee2019understanding,yadollahi2017current}. 
The core task of TER is to identify the specific emotion categories expressed in the text, such as happiness, sadness, anger, surprise, etc. \cite{chatterjee2019understanding}. Compared to textual sentiment analysis \cite{mohammad2016sentiment,zhang2018deep,nandwani2021review,liu2022sentiment} (e.g., positive, negative, neutral), 
TER offers a more fine-grained characterization of emotional semantics, capturing richer and more subtle emotional expressions in the text \cite{deng2021survey,peng2022survey}. 
Table \ref{diff e s} illustrates the main differences between TER and textual sentiment analysis.
In recent years, generative models have shown great potential in TER tasks. 
By learning the intrinsic patterns and distributions of textual data, generative models, especially LLMs or their predecessors, can generate emotionally expressive text, providing new perspectives and methods for emotion recognition \cite{zhou2018emotional}. 
Based on this, this section reviews a series of works of generative models in TER tasks.

\begin{table}[h]
\centering
\caption{Difference between Emotion Recognition and Sentiment Analysis.}
\label{diff e s}
\resizebox{\textwidth}{!}{%
\begin{tabular}{|c|c|c|}
\hline
           & Textual Emotion Recognition (TER)        & Textual Sentiment Analysis                                 \\ \hline
Type & Specific Emotion           & Sentiment Polarity                                 \\ \hline
Typical Categories   & Happiness, Sadness, Anger, Fear, Surprise, Disgust & Positive, Neutral, Negative  \\ \hline
\end{tabular}%
}
\end{table}

For instance, 
in terms of data augmentation, Nedilko \cite{nedilko2023generative} utilizes ChatGPT to augment the dataset by translating the data into different languages for the task of TER. Similarly, Koptyra et al. \cite{koptyra2023clarin} investigate the feasibility of employing ChatGPT for automatic emotional text generation and annotation in the context of TER.
Pico et al. \cite{pico2024exploring} study the application of text-generating LLMs for TER with the aim of generating more emotional knowledge, in the form of beliefs, that can be utilized by an emotional agent.

\begin{figure}[t]
  \centering
  \includegraphics[width=1.0\textwidth]{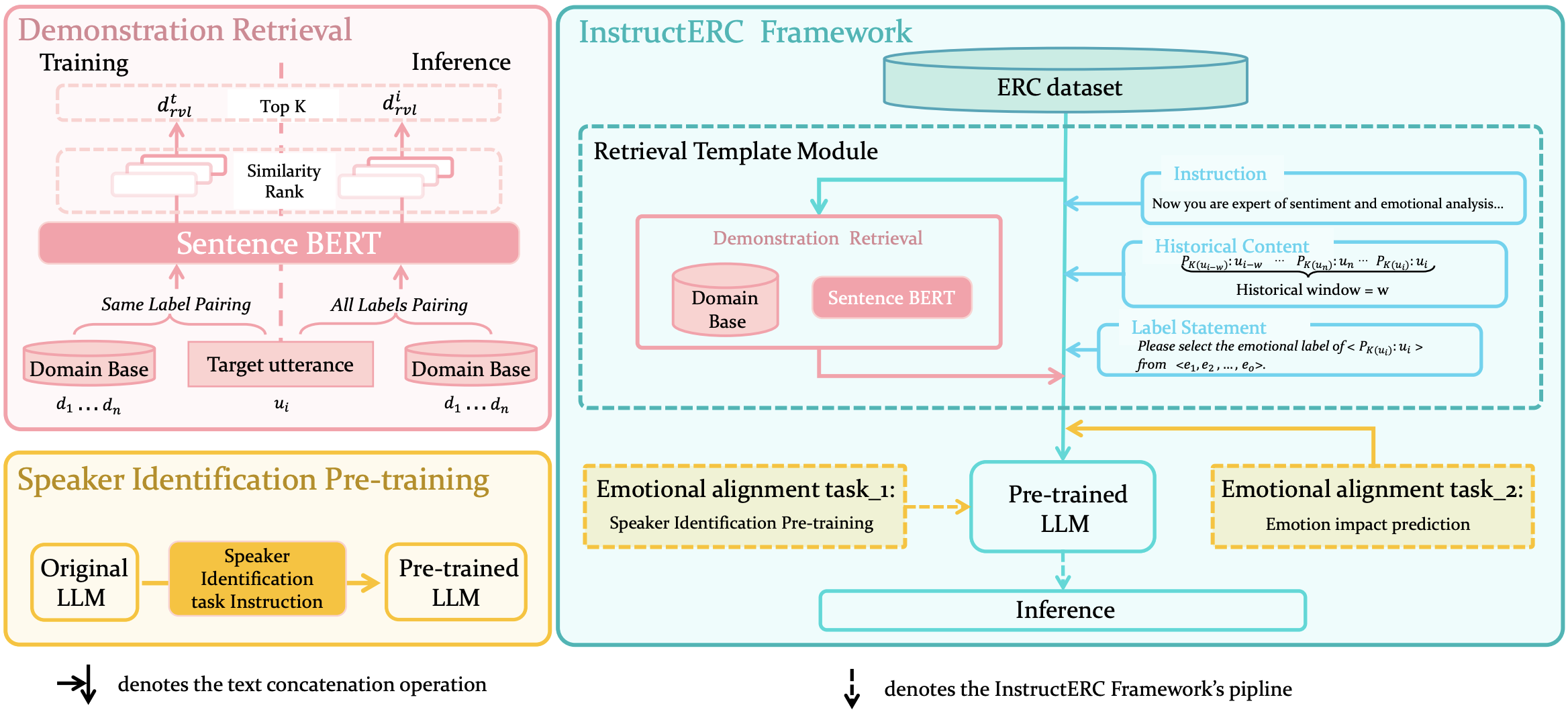}
  \caption{
  A LLM-based framework for TER proposed by \cite{lei2023instructerc}:  
It introduces an effective retrieval template module that explicitly integrates multi-granularity information. Moreover, it develops two additional emotion alignment tasks, which implicitly model the role relationships and future emotional tendencies.
  }
  \label{InstructERC}
\end{figure}

Furthermore, in terms of feature extraction,
Ghosal et al. \cite{ghosal2020cosmic} leverage COMET \cite{bosselut2019comet}, a GPT-based model, to extract emotional textual features, such as mental states, events, and causal relations, for the task of TER. These features are then incorporated into their proposed framework, COSMIC, which aims to enhance the understanding of emotional dynamics in conversations by capturing the underlying commonsense knowledge that influences the emotional states of the interlocutors.
Hama et al. \cite{hama2024emotion} demonstrate that the LLMs can achieve excellent performance for emotion labels with a limited number of training samples. Additionally, they introduce a multi-step prompting technique to further enhance the discriminative capacity of the LLMs for different emotion labels.
InstructERC \cite{lei2023instructerc} reformulates the TER task by transitioning from a discriminative framework to a generative framework based on LLMs,
as illustrated in Figure \ref{InstructERC}. CKERC \cite{fu2024ckerc} is a joint LLMs with commonsense knowledge framework for TER, which utilizes commonsense knowledge for LLM pre-training to finetune implicit clues information.

\section{Physiological Signal-based Emotion Recognition}
\label{ps}
\quad
In addition to the emotion recognition based on speech, facial images, and text mentioned above, physiological signals have garnered significant attention in the field of emotion recognition because they directly reflect an individual's bodily changes during emotional experiences. 
These changes are usually not influenced by an individual's subjective consciousness and can therefore provide an objective reflection of emotional states  \cite{egger2019emotion,shu2018review}. 
The physiological signals used in emotion recognition include EEG, ECG, HRV, EDA, and others, each with its own unique characteristics. 
For instance, EEG signals have the advantages of high temporal resolution and directly reflecting neural activity \cite{alarcao2017emotions}, enabling them to capture subtle changes in brain activity. 
As a result, EEG is widely used in emotion recognition research \cite{zhong2020eeg,li2017human}. 
On the other hand, ECG signals reflect changes in cardiac activity and are closely related to emotional states \cite{hasnul2021electrocardiogram,sarkar2020self}. HRV signals indicate the variability of heart rate, which can reflect the activity of the autonomic nervous system \cite{ferdinando2014emotion}. 
EDA signals reflect changes in skin conductance and are associated with emotional arousal \cite{shukla2019feature}. 
These signals provide valuable insights into the physiological changes associated with different emotional experiences \cite{shu2018review}. 
In recent years, generative models have shown great advantages in physiological signal-based emotion recognition \cite{qiu2023review}, similar to their applications in the fields of SER, FER, and TER. 
These models have demonstrated their potential in various aspects, such as data augmentation, feature extraction, and cross-domain, providing new impetus for the development of this field.
Table \ref{ps} presents the literature on generative models for physiological signal-based emotion recognition.

\begin{figure}[t]
   \centering
   \includegraphics[width=0.9\textwidth]{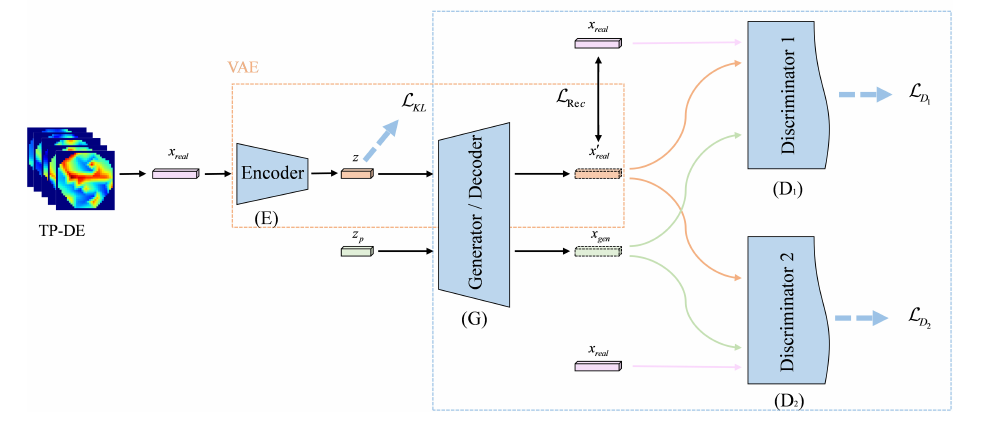}
   \caption{
   The framework of VAE-D2GAN model for data augmentation in physiological signal-based emotion recognition from \cite{Bao2021}: It consists of four parts: an encoder, a generator, and two discriminators. The encoder and generator form the VAE. The encoder maps real samples to latent vectors, which the generator then uses to create artificial samples. The generator and the two discriminators form the GAN. The generator learns the real data distribution from the gradients provided by the discriminators, which distinguish real samples from generated ones.
  }
   \label{eeg_gragh}
\end{figure}

Data augmentation is the primary application of generative models in physiological signal-based emotion recognition. Researchers conduct a series of studies in this direction. For example,  
Luo et al. \cite{luo2018eeg} propose a Conditional Wasserstein GAN (CWGAN) framework for EEG data augmentation in emotion recognition, which generates high-quality realistic-like EEG data in differential entropy (DE) form, judged by three indicators, to supplement the data manifold and improve emotion classification performance.
As shown in Figure \ref{eeg_gragh}, Bao et al. \cite{Bao2021} propose VAE-D2GAN, a data augmentation model for EEG emotion recognition, which combines a VAE model with a dual discriminator GAN to learn the distributions of DE features under five classical frequency bands and generate diverse artificial EEG samples for training.
Bhat and Hortal \cite{bhat2021gan} present a model based on Wasserstein GAN with gradient penalty, capable of generating new data features to augment the original dataset.
Zhang et al. \cite{zhang2022ganser} propose a GAN-based framework with self-supervised learning to generate EEG samples by learning an EEG generator, masking parts of EEG signals to synthesize potential signals, and utilizing masking possibility as prior knowledge to extract distinguishable features and generalize the classifier.
Zhang et al. \cite{zhang2024beyond} propose a emotional subspace constrained GAN model that addresses the limitations of existing GANs in generating reliable augmentations for under-represented minority emotions in imbalanced EEG datasets by introducing an EEG editing paradigm and incorporating diversity-aware and boundary-aware losses to constrain the augmented emotional subspace.
Similar works on EEG data augmentation based on GANs also include \cite{haradal2018biosignal,Pan2021,Luo2020,kalashami2022eeg}.
Apart from GAN-based models, other generative approaches have also been explored. Siddhad et al. \cite{siddhad2024enhancing} employ a conditional DM to generate raw synthetic EEG data, while Tosato et al. \cite{tosato2023eeg} and Yi et al. \cite{yi2024weighted} also utilize DMs for EEG data generation.

In addition to data augmentation, generative models can be employed for feature extraction in physiological signal-based emotion recognition, among which AE-based models are widely used.
For example,
Lan et al. \cite{lan2017unsupervised} use an AE model to learn subject-specific salient frequency components from EEG signals' power spectral density, automatically identifying the most discriminative features without predefined frequency band ranges.
Rajpoot et al. \cite{rajpoot2022subject} employ an unsupervised LSTM with channel-attention AEs to obtain subject-invariant latent vector subspace representations and a CNN with attention framework to perform EEG emotion recognition on the encoded latent space representations.
Similar AE-based methods also include \cite{jirayucharoensak2014eeg,li2019variational,bethge2022eeg2vec,liu2020eeg,qing2019interpretable,li2020latent}.
In addition, some researchers attempt to combine GAN-based approaches. 
For example, Gu et al. \cite{Gu2023} integrates generative adversarial learning into a hybrid model of Graph Convolutional Neural Network (GCNN) and LSTM, which utilizes a GAN to generate latent representations of EEG signals, aiming to extract more discriminative features and improve classification performance.

To effectively utilize both unlabeled and labeled data, semi-supervised learning approaches are explored. Notably, Zhang et al. \cite{zhang2021deep} introduced a semi-supervised framework for EEG emotion recognition using an attention-based recurrent AE model. Their approach consists of two components: an unsupervised component that maximizes the consistency between the original and reconstructed input data, and a supervised component that minimizes the cross-entropy between the input and output labels.
To address the cross-domain problem, researchers propose a series of methods. For example, 
Chai et al. \cite{chai2016unsupervised} introduce a subspace-aligned AE framework to obtain a universal representation across training and testing domains for EEG emotion recognition. 
Wang et al. \cite{wang2022multi} propose a multimodal VAE to utilize the multi-modal data (EEG and eye movement data) from the source domain to assist in EEG emotion recognition in the target domain.
Huang et al. \cite{huang2022generator} propose a GAN-based domain adaptation method with a knowledge-free mechanism, which converts the source domain in the data space into the target domain through a novel EEG content loss function.

\begin{landscape}
\begin{table}[]
\centering
\caption{Literature on Generative Models for Physiological Signal-based Emotion Recognition.}
\label{ps}
\resizebox{\columnwidth}{!}{%
\begin{tabular}{|l|l|l|l|l|l|l|}
\hline
Reference                                  & Year & Model           & Application              & Modality & Dataset      & Performance (\%) \\ \hline
Luo et al. \cite{luo2018eeg}               & 2018 & CWGAN           & Data Augmentation        & EEG      & SEED         & ACC: 86.96       \\ \hline
Bao et al. \cite{Bao2021}                  & 2021 & VAE + GAN       & Data Augmentation        & EEG      & SEED/SEED-IV & ACC: 92.5/82.3   \\ \hline
Bhat and Hortal \cite{bhat2021gan}         & 2021 & Wasserstein GAN & Data Augmentation        & EEG      & DEAP         & ACC: 71.50       \\ \hline
Zhang et al. \cite{zhang2022ganser}                     & 2022 & GAN & Data Augmentation        & EEG & DEAP/SEED/DREAMER & ACC: 94.38/97.71/85.28 \\ \hline
Zhang et al. \cite{zhang2024beyond}        & 2024 & GAN             & Data Augmentation        & EEG      & DEAP/SEED    & ACC: 96.68/97.14 \\ \hline
Haradal et al. \cite{haradal2018biosignal} & 2018 & GAN             & Data Augmentation        & ECG      & ECG 200      & ACC: 76.20       \\ \hline
Pan and Zheng \cite{Pan2021}               & 2021 & GAN             & Data Augmentation        & EEG      & DEAP         & ACC: 90.26       \\ \hline
Luo et al. \cite{Luo2020}                  & 2020 & GAN             & Data Augmentation        & EEG      & SEED/DEAP    & ACC: 67.7/50.8   \\ \hline
Kalashami et al. \cite{kalashami2022eeg}   & 2022 & CWGAN           & Data Augmentation        & EEG      & DEAP         & ACC: 71.9        \\ \hline
Siddhad et al. \cite{siddhad2024enhancing} & 2024 & DM              & Data Augmentation        & EEG      & DEAP         & ACC: 71.50       \\ \hline
Tosato et al. \cite{tosato2023eeg}         & 2023 & DM              & Data Augmentation        & EEG      & SEED         & ACC: 92.98       \\ \hline
Lan et al. \cite{lan2017unsupervised}      & 2017 & AE              & Feature Extraction       & EEG      & SEED         & ACC: 59.03       \\ \hline
Rajpoot et al. \cite{rajpoot2022subject}                & 2022 & AE  & Feature Extraction       & EEG & DEAP/SEED/CHB-MIT & ACC: 69.5/76.7/72.3    \\ \hline
Jiarayucharoensak et al. \cite{jirayucharoensak2014eeg} & 2014 & SAE & Feature Extraction       & EEG & DEAP              & ACC: 49.52             \\ \hline
Li et al. \cite{li2019variational}         & 2019 & VAE             & Feature Extraction       & EEG      & DEAP/SEED    & ACC: 72.43/84.29 \\ \hline
Bethge et al. \cite{bethge2022eeg2vec}     & 2022 & CAE             & Feature Extraction       & EEG      & SEED         & ACC: 69          \\ \hline
Liu et al. \cite{liu2020eeg}               & 2020 & SAE             & Feature Extraction       & EEG      & DEAP/SEED    & ACC: 92.86/96.77 \\ \hline
Qing et al. \cite{qing2019interpretable}                & 2019 & SAE & Feature Extraction       & EEG & DEAP/SEED         & ACC: 63.09/75          \\ \hline
Li et al. \cite{li2020latent}              & 2020 & VAE             & Feature Extraction       & EEG      & DEAP/SEED    & ACC: 72.43/84.29 \\ \hline
Gu et al. \cite{Gu2023}                    & 2023 & GAN             & Feature Extraction       & EEG      & DEAP/SEED    & ACC: 94.78/97.28 \\ \hline
Zhang et al. \cite{zhang2021deep}          & 2021 & AE              & Semi-supervised Learning & EEG      & SEED         & ACC: 91.17       \\ \hline
Chai et al. \cite{chai2016unsupervised}    & 2016 & AE              & Cross-domain & EEG      & SEED         & ACC: 62.50       \\ \hline
Wang et al. \cite{wang2022multi}                        & 2022 & VAE & Cross-domain & EEG & SEED/SEED-IV      & ACC: 89.64/73.82       \\ \hline
Huang et al. \cite{huang2022generator}     & 2022 & AE              & Cross-domain & EEG      & DEAP         & ACC: 63.85       \\ \hline
\end{tabular}%
}
\end{table}
\end{landscape}

\section{Multimodal Emotion Recognition (MER)}
\label{multi er}
\quad
Unlike unimodal emotion recognition mentioned in Sections \ref{ser} - \ref{ps} above, Multimodal Emotion Recognition (MER) is a research field that aims to identify emotional states by combing the information from different modalities, including speech, facial images, text, EEG, eye movements, and functional magnetic resonance imaging (fMRI), and so on \cite{poria2017review,ezzameli2023emotion,tzirakis2017end,zhang2020emotion}. 
By integrating these heterogeneous data, MER can more comprehensively capture and analyze subtle differences in emotional expressions, thereby significantly improving the performance of emotion recognition.
In recent years, generative models have been increasingly applied in the field of MER, showing potential in various aspects such as data augmentation, feature extraction, semi-supervised learning, and modality completion.
Table \ref{multi} summarizes a series of representative works on generative models for MER.

Similar to unimodal emotion recognition, the primary application of generative models in MER is also data augmentation.
For example, 
Ma et al. \cite{ma2022data} design a multimodal CGAN based on Hirschfeld--Gebelein--R$\acute{\textup{e}}$nyi (HGR) maximal correlation~\cite{hirschfeld1935connection,gebelein1941statistische,renyi1959measures} to generate audio-visual data with different emotion categories.
Luo et al. \cite{luo2019gan} implement a conditional boundary Equilibrium GAN \cite{berthelot2017began} to generate artificial differential entropy features for EEG data, eye movement data, and their direct connections.
Yan et al. \cite{yan2021simplifying} propose a CGAN-based framework that uses eye movement signals as the sole input to map information onto EEG-eye movement features.
Chao et al. \cite{chao2018generating} employ a GAN framework with a cyclic loss to learn the bi-directional generative relationship between acoustic features and fMRI signals, enabling the derivation of fMRI-enriched acoustic features for emotion classification.

In addition to data augmentation, feature extraction is an important aspect of MER. 
For example, 
Yan et al. \cite{yan2017identifying} propose a method that uses a bimodal deep AE and a fuzzy integral-based technique \cite{1572261549724574208} to extract high-level fused features from EEG signals and eye movements, which are then input into a SVM classifier framework to obtain an emotion classification model.
Similar works based on the bimodal deep AE also include \cite{guo2019multimodal,zhang2020expression,shixin2022autoencoder,hamieh2021multi}.
Nguyen et al. \cite{nguyen2021deep} propose a MER model that combines a two-stream AE with a LSTM network to perform end-to-end trainable joint learning of temporal and compact-representative features from visual and audio signals.
Ma et al. \cite{ma2019end} propose a deep framework with CNN and LSTM for MER that learns feature mappings from multimodal data by introducing a correlation loss based on HGR maximal correlation and a reconstruction loss of AEs to capture common and private information among different modalities, respectively.
Zheng et al. \cite{zheng2022multi} propose a novel Multi-channel Weight-sharing AE model with a convolutional encoder network for improved feature extraction and a scalable feature fusion method based on multi-head attention mechanism (CMHA) to fuse the output features of multiple encoders for MER.

\begin{figure}[t]
   \centering
   \includegraphics[width=0.9\textwidth]{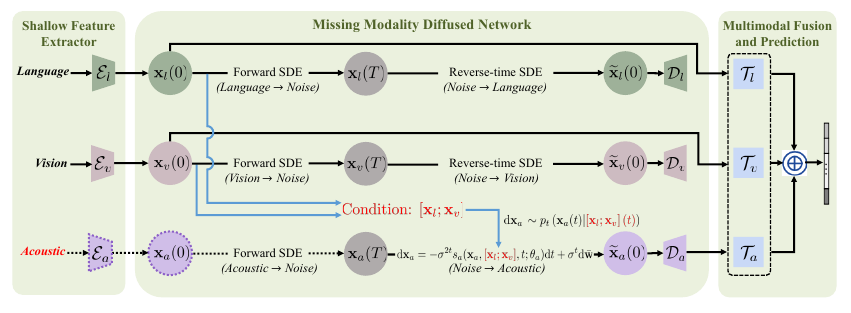}
   \caption{The IMDer framework for modality completion in MER from \cite{wang2024incomplete}: It handles incomplete data (e.g., missing acoustic modality) by extracting shallow features from the available modalities. Then the diffused network samples the missing modality data from a noise distribution and then recovers it through a reverse diffusion process. Finally, the recovered data is combined with the other modality data for multimodal fusion and prediction to regress emotional labels.}
   \label{multi graph}
 \end{figure}

Semi-supervised learning is another aspect where generative models are considered in MER. For example, 
Du et al. \cite{du2017semi} propose a semi-supervised MER framework based on a multi-view VAE that imposes a mixture of Gaussians assumption on the posterior approximation of latent variables. This framework can leverage both labeled and unlabeled samples from multiple modalities and automatically learn the weight factor for each modality.
Liang et al. \cite{liang2019semi} propose a semi-supervised learning approach for MER based on an improved GAN, CT-GAN \cite{wei2018improving}, which utilizes unlabeled data from acoustic and visual modalities.

In addition to the aforementioned aspects, 
another direction that requires attention in MER is modality completion \cite{jaques2017multimodal,geetha2024multimodal}. 
It refers to situations where some modalities in multimodal data are missing or incomplete due to reasons such as device failure, signal loss, or data corruption. 
By completing the missing modalities, incomplete multimodal data can be fully utilized to ensure the performance of emotion recognition. 
Generative models provide an effective approach for modal completion. 
For example, 
Liu et al. \cite{liu2024contrastive} employ multiple AEs to generate high-quality missing modality data by reconstructing it from the available modality features and the invariant features learned through contrastive learning. 
As shown in Figure \ref{multi graph}, Wang et al. \cite{wang2024incomplete} propose an Incomplete Multimodality-Diffused emotion recognition (IMDer) method that exploits a score-based DM to recover missing modalities while maintaining distribution consistency and semantic disambiguation by using available modalities as a condition to guide the diffusion-based recovering process.

\begin{landscape}
\begin{table}[]
\centering
\caption{Literature on Generative Models for MER.}
\label{multi}
\resizebox{\columnwidth}{!}{%
\begin{tabular}{|l|l|l|l|l|l|l|}
\hline
Reference                             & Year & Model & Application              & Modalities                       & Dataset            & Performance (\%) \\ \hline
Ma et al. \cite{ma2022data}              & 2022 & CGAN & Data Augmentation   & Speech + Image        & eNTERFACE’05/RAVDESS/CMEW      & ACC: 49.48/65.90/46.19 \\ \hline
Luo et al. \cite{luo2019gan}          & 2019 & CGAN  & Data Augmentation        & EEG + Eye Movement               & SEED/SEED-V        & ACC: 90.33/68.32 \\ \hline
Yan et al. \cite{yan2021simplifying}     & 2021 & CGAN & Data Augmentation   & EEG + Eye Movement    & SEED/SEED-IV/SEED-V            & ACC: 93.05/86.55/80.37 \\ \hline
Chao et al. \cite{chao2018generating}    & 2018 & GAN  & Data Augmentation   & Speech + fMRI         & AudiofMRI Parallel Set/IEMOCAP & ACC: 49.58             \\ \hline
Yan et al. \cite{yan2017identifying}  & 2017 & AE    & Feature Extraction       & EEG + Eye Movement               & Self               & ACC: 64.26       \\ \hline
Guo et al. \cite{guo2019multimodal}   & 2019 & AE    & Feature Extraction       & Eye movements + Eye images + EEG & SEED V             & ACC: 79.63       \\ \hline
Zhang \cite{zhang2020expression}      & 2020 & DAE   & Feature Extraction       & Image + EEG                      & DEAP               & ACC: 85.71       \\ \hline
Peng et al. \cite{shixin2022autoencoder} & 2022 & AE   & Feature Extraction  & Speech + Text         & IEMOCAP/MELD/CMU-MOSI          & ACC: 74.8/63.64/79.85  \\ \hline
Hamieh et al. \cite{hamieh2021multi}  & 2021 & AE    & Feature Extraction       & Speech + Image + Text            & UlmTSST            & ACC: 72.95       \\ \hline
Nguyen et al. \cite{nguyen2021deep}   & 2021 & AE    & Feature Extraction       & Speech + Image                   & RECOLA             & ACC: 47.4        \\ \hline
Ma et al. \cite{ma2019end}            & 2019 & AE    & Feature Extraction       & Speech + Image                   & eNTERFACE’05       & ACC: 85.43       \\ \hline
Zheng et al. \cite{zheng2022multi}       & 2022 & AE   & Feature Extraction  & Speech + Image + Text & IEMOCAP/MSP-IMPROV             & ACC: 85.5/70.9         \\ \hline
Du et al.\cite{du2017semi}            & 2017 & VAE   & Semi-supervised Learning & EEG + Eye Movement               & DEAP/SEED          & ACC: 45.6/96.8   \\ \hline
Liang et al. \cite{liang2019semi}     & 2019 & GAN   & Semi-supervised Learning & Speech + Image                   & IEMOCAP            & UAR: 63.98       \\ \hline
Liu et al. \cite{liu2024contrastive}     & 2024 & AE   & Modality Completion & Speech + Image + Text & IEMOCAP/MSP-IMPROV             & ACC: 65.13/62.43       \\ \hline
Wang et al. \cite{wang2024incomplete} & 2024 & DM    & Modality Completion      & Speech + Image                   & CMU-MOSI/CMU-MOSEI & ACC: 79.6/80.9   \\ \hline
\end{tabular}%
}
\end{table}
\end{landscape}

\section{Discussion}
\label{discussion}
\subsection{Summary Findings}
\begin{figure}
  \centering
  \begin{subfigure}[b]{0.5\textwidth}
    \centering
    \includegraphics[width=\textwidth]{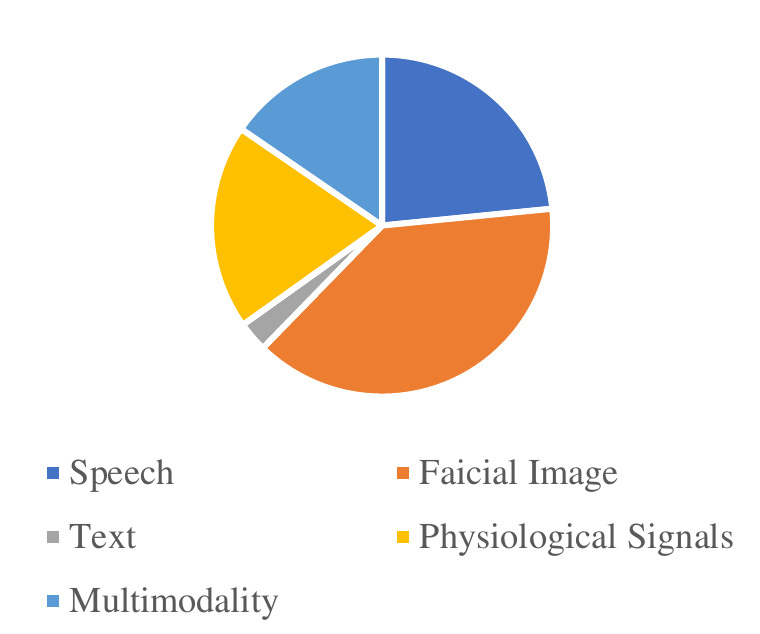}
    \caption{}
    \label{paper}
  \end{subfigure}%
  \begin{subfigure}[b]{0.5\textwidth}
    \centering
    \includegraphics[width=\textwidth]{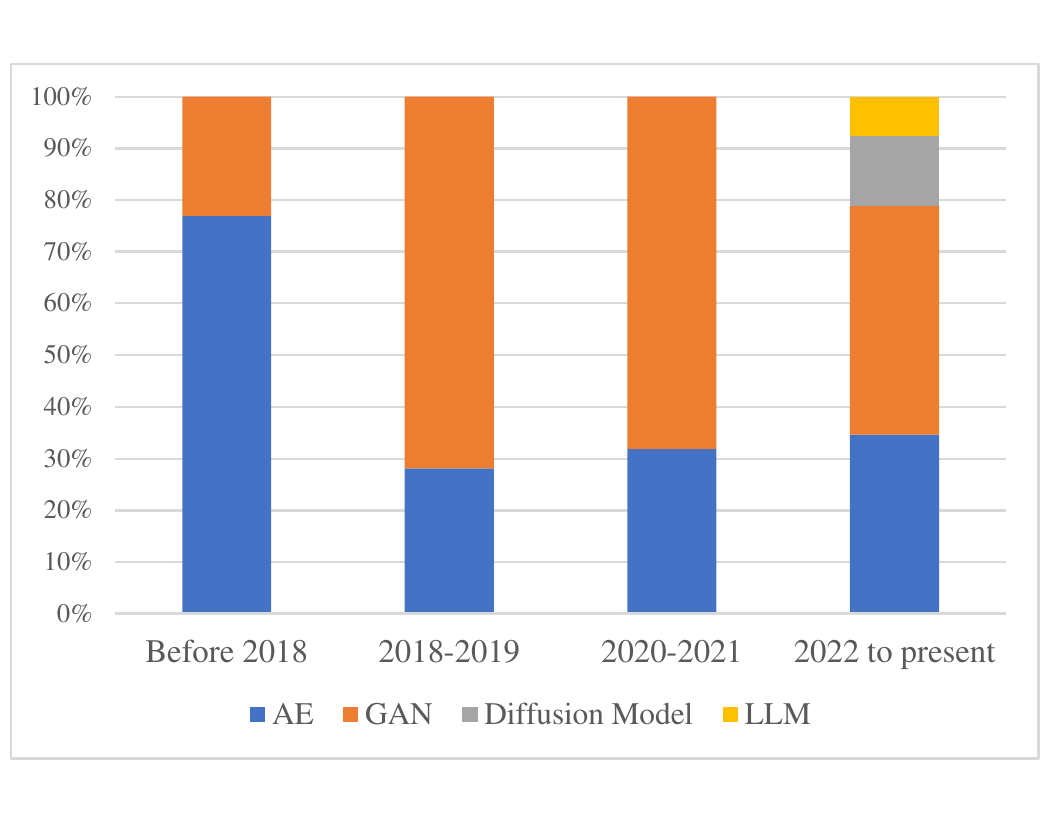}
    \caption{}
    \label{model}
  \end{subfigure}
  \caption{(a) The proportion of different modalities being used, (b) The proportion of different generative models used over time.}
  \label{fig:combined}
\end{figure}

\quad
Firstly, we conduct a statistical analysis of the above research papers, as shown in Figure \ref{fig:combined}, which includes two subfigures \ref{paper} and \ref{model}. From subfigure \ref{paper}, it can be seen that among various modalities, facial images are the most commonly used by generative models for emotion recognition, reflecting that facial information can express emotions more intuitively and richly. Next are speech signals, while physiological signals, multimodal signals, and text modalities are less common. Subfigure \ref{model} shows the usage trends of various generative models in emotion recognition research during different periods. 
It can be seen that AEs are most common initially, and over time, the use of GANs is increased. Recently, DMs and LLMs have also begun to gradually appear in research. 
This evolution reflects the progress of generative technology, with new and more complex models continuously being developed and applied to the field of emotion recognition.

Secondly, generative models have applications in various aspects of emotion recognition, including data augmentation, feature extraction, semi-supervised learning, cross-domain, etc., among which data augmentation and feature extraction are the most common. This may be because the acquisition and annotation of emotion data is quite expensive, especially for facial images and speech data. Utilizing generative models, more samples can be synthesized from existing data to expand the training set. 
At the same time, the feature representations learned by generative models can also be easily used for downstream emotion recognition tasks to improve performance.

Thirdly, although there are common integration points of generative models in emotion recognition based on different modalities, the specific usage still needs to consider the characteristics of each modality. 
For example, when generative models are used for data augmentation in SER, the temporal information of speech data can be preserved in the feature space or sample space, while in FER, it almost always operates directly in the sample space to generate image data. 
Meanwhile, in terms of enhancing robustness, SER pays attention to adversarial sample generation, while FER conducts more research directly from the perspective of dealing with noisy environments such as occlusion and complex backgrounds during sample acquisition. 
These differences indicate that applying generative models to emotion recognition based on different modalities requires fully considering the unique attributes of their data and designing and optimizing models in a targeted manner.

Fourthly, it can be seen that among various emotion recognition models, those based on AE and GAN are used the most, possibly because they appeared earlier, and both theory and practice are relatively mature, with widespread applications. 
The use of DM and LLM is relatively less, possibly because as of now, the application of DM in emotion recognition is still in its infancy, while LLM is mainly used with text modalities. 
However, due to their excellent generative capabilities and ability to model prior knowledge, it is believed that they will play a greater role in the field of emotion recognition in the future.

Fifthly, it is worth mentioning that we also find relatively few literature on generative models in TER, which may be due to the fact that there has been less mining of fine-grained emotional attributes in text in the past. 
However, we also note that the application of LLM in TER is increasing. Its powerful semantic understanding and generative capabilities are expected to promote the further development of TER.

Finally, it can be seen that eye movement signals are often combined with other modalities for emotion recognition. However, since the ability of eye movement signals alone to comprehensively represent emotions is limited, they are usually not used independently for emotion recognition, but rather integrated with other modalities to improve overall performance, indicating that multimodal fusion is a major trend in emotion recognition.

\subsection{Future Prospects}
\quad
In the above discussion, we explore various applications of generative models in the field of emotion recognition. Looking forward, research on using generative models for emotion recognition remains full of opportunities, including the following aspects:

Firstly, an exciting development direction is to combine advanced generative models such as DMs with Transformer architectures \cite{peebles2023scalable,bao2023all}. 
DMs excel at generating high-quality, diverse data samples and have shown more remarkable results than AEs and GANs in an increasing number of domains \cite{dhariwal2021diffusion,stypulkowski2024diffused}. 
On the other hand, Transformer architectures, with their powerful sequence modeling capabilities and strong contextual modeling abilities, have brought about a paradigm shift \cite{vaswani2017attention}. 
Combining the two can produce a more powerful generative model capable of simultaneously processing multimodal data such as images, speech, and text, thereby capturing multiple aspects of human emotions more comprehensively and accurately. This combination can significantly enhance the performance of emotion recognition, enabling systems to understand more subtle and complex emotional expressions.

Secondly, combining generative models with other cutting-edge AI technologies such as reinforcement learning \cite{arulkumaran2017deep} and federated learning \cite{li2020review,zhang2021survey} can further push the boundaries of emotion recognition. Reinforcement learning specializes in modeling and optimizing multi-stage decision-making processes, and emotion recognition and generative models are often dynamic processes involving multiple steps such as environment perception, strategy determination, and performance evaluation. 
Combining the two can achieve more interactive and adaptive emotional systems \cite{bilquise2022emotionally}. Furthermore, due to the privacy nature of emotional data, it is often difficult to collect on a large scale, which limits the improvement of model performance. Federated learning can allow distributed data to contribute to model training while protecting privacy \cite{latif2020federated,zhao2023privacy}. By training models on personal devices and then only uploading parameter updates instead of raw data, it is hoped that more powerful and personalized emotion models can be obtained.

Thirdly, with the continuous development of generative models, their applications in emotion recognition are expanding to new scenarios such as virtual reality (VR) and augmented reality (AR) \cite{marin2020emotion,papoutsi2021virtual}. In these immersive environments, user's emotions and experiences are closely linked. 
For example, in VR games or social interactions, generating responsive scenes, characters, and dialogues in real-time based on the user's emotional state can create a more immersive experience \cite{felnhofer2015virtual}. 
In AR navigation or shopping, overlaying emotion-related information and recommendations can provide more thoughtful services \cite{valente2022empathic}. 
Since VR/AR devices can collect multimodal user data such as gaze and actions, they provide richer clues for emotion recognition. Generative models can leverage this data to create more diverse and dynamic emotional responses.

Finally, this survey mainly discusses using generative models to assist emotion recognition, but another promising direction is to directly synthesize emotionally rich content, such as emotional speech, text, images, and videos \cite{wang2023emotional,lian2023gpt}. 
This has important application value in many fields. 
For example, in film and animation production, automatically generating dubbing and background music with specific emotional tones can reduce the cost of manual recording \cite{ji2024muser}. 
In intelligent customer service systems, empathetic and personalized responses can be synthesized in real-time based on the conversation content and user emotions, improving user satisfaction \cite{leocadio2024customer}. 
In the field of education, emotionally rich learning materials (such as stories and dialogues) can be automatically generated to create a more vivid and interesting learning experience \cite{hong2023visual}. 
This puts forward higher requirements for the design of generative models and requires more refined conditional control mechanisms. 
A beneficial prior work is the data augmentation discussed in this survey, i.e., using generative models to synthesize samples with different emotions to expand the dataset. 
In the future, data augmentation is expected to provide a data foundation and prior knowledge for further high-quality emotional content generation.

\section{Conclusion}
\label{conclusion}
\quad

In this paper, we conduct the first systematic survey of the progress of generative technology in the field of human emotion recognition. 
By analyzing over 320 relevant papers, we delve into the extensive impact of generative models on various aspects and modalities of emotion recognition. 
This work fills the gap in existing reviews and provides valuable references for researchers in this field.

In Section \ref{intro}, we introduce the background knowledge of emotion recognition and affective computing, clarifying the close relationship between the two. 
Emotion recognition is one of the key technologies for realizing affective computing and is crucial for building intelligent, natural, and friendly human-computer interaction systems. 
Generative models, with their powerful data modeling and generation capabilities, provide new ideas for emotion recognition in multiple aspects. 
Based on this, we propose the taxonomy for the application of generative models in emotion recognition and clarify the main contributions of this review, which is to systematically summarize the current state of generative models in emotion recognition and to outlook future research directions.

In Section \ref{diff}, we make a detailed comparison between this review and other relevant reviews. 
Although there have been some reviews on emotion recognition or generative models, no work has systematically discussed the combination of the two. This review fills this gap and helps researchers comprehensively understand the latest developments in this interdisciplinary field. 
In Section \ref{gm}, in order to help readers better understand the subsequent content, we briefly introduce the mathematical principles of several common generative models, including AE, GAN, DM, and LLM. 
These models have their own characteristics and play important roles in different tasks of emotion recognition. Section \ref{dataset} outlines the commonly used datasets for generative models in emotion recognition tasks, covering multiple modalities such as speech, facial images, text, and physiological signals.

Then, from multiple perspectives such as data augmentation, feature extraction, semi-supervised learning, and cross-domain, we elaborate on the applications of generative models in emotion recognition based on different modalities, including SER in Section \ref{ser}, FER in Section \ref{fer}, TER in Section \ref{ter}, physiological signal-based emotion recognition in Section \ref{ps}, and MER in Section \ref{multi er}. 
Through comprehensive analysis, we demonstrate that generative models can (1) generate high-quality emotional data to alleviate the problem of data scarcity, (2) learn high-level feature representations of emotional data to enhance the discriminative power of features, (3) utilize unlabeled data for semi-supervised learning to reduce the dependence on large amounts of labeled data, (4) generate cross-domain data to enhance the generalization ability of models, and other aspects. 
These applications greatly promote the development of emotion recognition. 
In Section \ref{discussion}, we analyze the characteristics of different generative models in emotion recognition based on different modalities, and emphasize the necessity of continuous innovation of generative models to further enhance their role in affective computing in the future.

In summary, generative models have brought new breakthroughs to emotion recognition, greatly expanding the depth and breadth of research. 
However, how to further tap the potential of generative models and build more intelligent, natural, and humanized affective computing systems still requires the joint efforts of academia and industry. 
We believe that with the continuous development of artificial intelligence technology, especially the increasing maturity of generative models, 
emotion recognition and the entire field of affective computing will usher in a broader development prospect and bring a better future for human-computer interaction. This review provides useful references and inspiration for achieving this goal.

\bibliographystyle{elsarticle-num} 
\bibliography{sample}






\end{document}